\pdfoutput=1

\documentclass[11pt]{article}

 \usepackage[]{acl}

\usepackage{times}
\usepackage{latexsym}

\usepackage[T1]{fontenc}

\usepackage[utf8]{inputenc}

\usepackage{microtype}

\usepackage{inconsolata}

\usepackage{xcolor}

\usepackage{dsfont}
\definecolor{mygreen}{RGB}{0,128,0}

\usepackage{multirow}
\usepackage{amsmath,bm}
\usepackage{amssymb}  
\usepackage{amsfonts}
\usepackage{graphicx} 
\usepackage[ruled]{algorithm2e} 

\SetKwInput{KwInput}{Input}                
\SetKwInput{KwOutput}{Output}              

\title{RobustSentEmbed: Robust Sentence Embeddings Using Adversarial Self-Supervised  Contrastive  Learning}

\author{Javad Rafiei Asl,\textsuperscript{1} Prajwal Panzade,\textsuperscript{1} Eduardo Blanco,\textsuperscript{2} Daniel Takabi,\textsuperscript{3} Zhipeng Cai\textsuperscript{1} \\
         \textsuperscript{1}Georgia State University, \textsuperscript{2}University of Arizona, \textsuperscript{3}Old Dominion University \\  jasl1@student.gsu.edu, eduardoblanco@arizona.edu, takabi@odu.edu }

\begin{document}
\maketitle
\begin{abstract}
Pre-trained language models (PLMs) have consistently demonstrated outstanding performance across a diverse spectrum of natural language processing tasks. Nevertheless, despite their success with unseen data, current PLM-based representations often exhibit poor robustness in adversarial settings. In this paper, we introduce RobustSentEmbed, a self-supervised sentence embedding framework designed to improve both generalization and robustness in diverse text representation tasks and against a diverse set of adversarial attacks. Through the generation of high-risk adversarial perturbations and their utilization in a novel objective function, RobustSentEmbed adeptly learns high-quality and robust sentence embeddings. Our experiments confirm the superiority of RobustSentEmbed over state-of-the-art representations. Specifically, Our framework achieves a significant reduction in the success rate of various adversarial attacks, notably reducing the BERTAttack success rate by almost half (from 75.51\% to 38.81\%). The framework also yields improvements of 1.59\% and 0.23\% in semantic textual similarity tasks and various transfer tasks, respectively.
\end{abstract}



\section{Introduction}
Pre-trained Language Models (PLMs) have demonstrated state-of-the-art performance in learning contextual word embeddings \cite{devlin_etal_2019_ber}, contributing to significant advancements in various Natural Language Processing (NLP) tasks \cite{yang2019xlnet, he2020deberta, ding2023parameter}. PLMs, including prominent models like BERT \cite{devlin_etal_2019_ber} and GPT-3 \cite{brown2020language}, have revolutionized text classification, sentence representation, and machine translation among a plethora of diverse NLP tasks. While PLMs have expanded their focus to include universal sentence embeddings, which effectively capture the semantic representation of input text, PLM-based sentence representations lack two crucial characteristics: generalization and robustness.


Extensive research efforts have been dedicated to the development of universal sentence embeddings employing PLMs \cite{reimers2019sentence, zhang2020unsupervised, neelakantan2022text, wang2023clsep}. Although these embeddings have demonstrated proficiency in generalization across various downstream tasks \cite{sun2019fine, gao2021simcse}, they exhibit limitations when subjected to adversarial settings and remain vulnerable to adversarial attacks \cite{nie2019adversarial, wang2021cline}. Existing research has highlighted the limited robustness of PLM-based representations \cite{garg2020bae, wu2023prada, hauser2023bert}. The vulnerability arises when these representations can be easily deceived by making small, imperceptible modifications to the input text. 


To address these limitations, we propose a method to obtain robust sentence embeddings called RobustSentEmbed. The main idea is to generate small adversarial perturbations and employ an efficient contrastive objective \cite{chen2020simple}. The goal is to enhance the adversarial resilience of the sentence embeddings. Specifically, our framework involves an iterative collaboration between an adversarial perturbation generator and the PLM-based encoder to generate high-risk perturbations in both token-level and sentence-level embedding spaces. RobustSentEmbed then employs a contrastive learning objective in conjunction with a token replacement detection objective to maximize the similarity between the embedding of the original sentence and the adversarial embedding of a positive pair (the former objective) as well as its edited sentence (the latter objective).


%

We have conducted comprehensive experiments to substantiate the efficacy of the RobustSentEmbed framework. The tasks encompass TextAttack \cite{morris2020textattack} assessments, adversarial Semantic Textual Similarity (STS) tasks, Non-adversarial STS tasks \cite{conneau2018senteval}, and transfer tasks \cite{conneau2018senteval}. Two initial series of experiments were designed to evaluate the robustness of our sentence embeddings against various adversarial attacks and tasks. Subsequently, we conducted two final series of experiments to assess the quality of our embeddings in the contexts of semantic similarity and natural language understanding. RobustSentEmbed demonstrates significant improvements in robustness, reducing the attack success rate from 75.51\% to 38.81\% against the  BERTAttack attack and from 71.86\% to 12.80\% on adversarial STS. Moreover, the framework outperforms existing methods in ten out of thirteen tasks while obtaining comparable results with the other three, showcasing improvements of 1.59\% and 0.23\% on STS tasks and NLP transfer tasks, respectively.



\textbf{Contributions.} Our main contributions are summarized as follows:
\begin{itemize}
  \item We introduce RobustSentEmbed, an innovative framework designed for generating sentence embeddings that are robust against adversarial attacks. Existing methods are vulnerable to such adversarial challenges. RobustSentEmbed fills this gap by generating high-risk perturbations and utilizing an efficient adversarial objective function.\footnote{Our code are publicly available at \url{https://github.com/jasl1/RobustSentEmbed}}
  \item We conduct comprehensive experiments to empirically evaluate the effectiveness of the RobustSentEmbed framework. The empirical findings substantiate the efficacy of our framework, as demonstrated by its superior performance in both robustness and generalization benchmarks.

\end{itemize}

%

\section{Related Work}
Recently, self-supervised methods using contrastive objectives have become prominent for learning effective and robust text representations: SimCSE, as outlined by \citet{gao2021simcse}, introduced a minimal augmentation method involving the application of two distinct dropout masks to predict the input sentence. The ConSERT model \cite{yan2021consert} employed four unique data augmentation techniques, namely adversarial attacks, token shuffling, cut-off, and dropout, to generate a variety of perspectives in order to carry out a contrastive objective. \citet{miao2021simple} utilized adversarial training to improve the robustness of contrastive learning. They achieved this by incorporating regularization into their learning objective, combining benign contrastive learning with an adversarial contrastive scenario. \citet{rima2023adversarial} proposed a novel method for training language processing models, combining adversarial training and contrastive learning. Their approach incorporates linear perturbations to input embeddings and uses contrastive learning to minimize the distance between the original and perturbed representations. \citet{pan2022improved} introduced a simple technique to improve the fine-tuning of Transformer-based encoders. Their method involves regularization by generating adversarial examples through word embedding perturbations and using contrastive learning to obtain noise-invariant representations.

Unlike existing approaches for training text representation through contrastive adversarial learning \cite{yan2021consert, miao2021simple, rima2023adversarial, pan2022improved}, our framework generates more efficient, high-risk perturbations at both the token-level and sentence-level within the embedding space. Furthermore, our framework utilizes a robust contrastive objective and incorporates an adversarial replaced token detection method, leading to high-quality text representations that yield improved generalization and robustness characteristics. 



\begin{figure*}
    \centering
    \includegraphics[width=\linewidth]{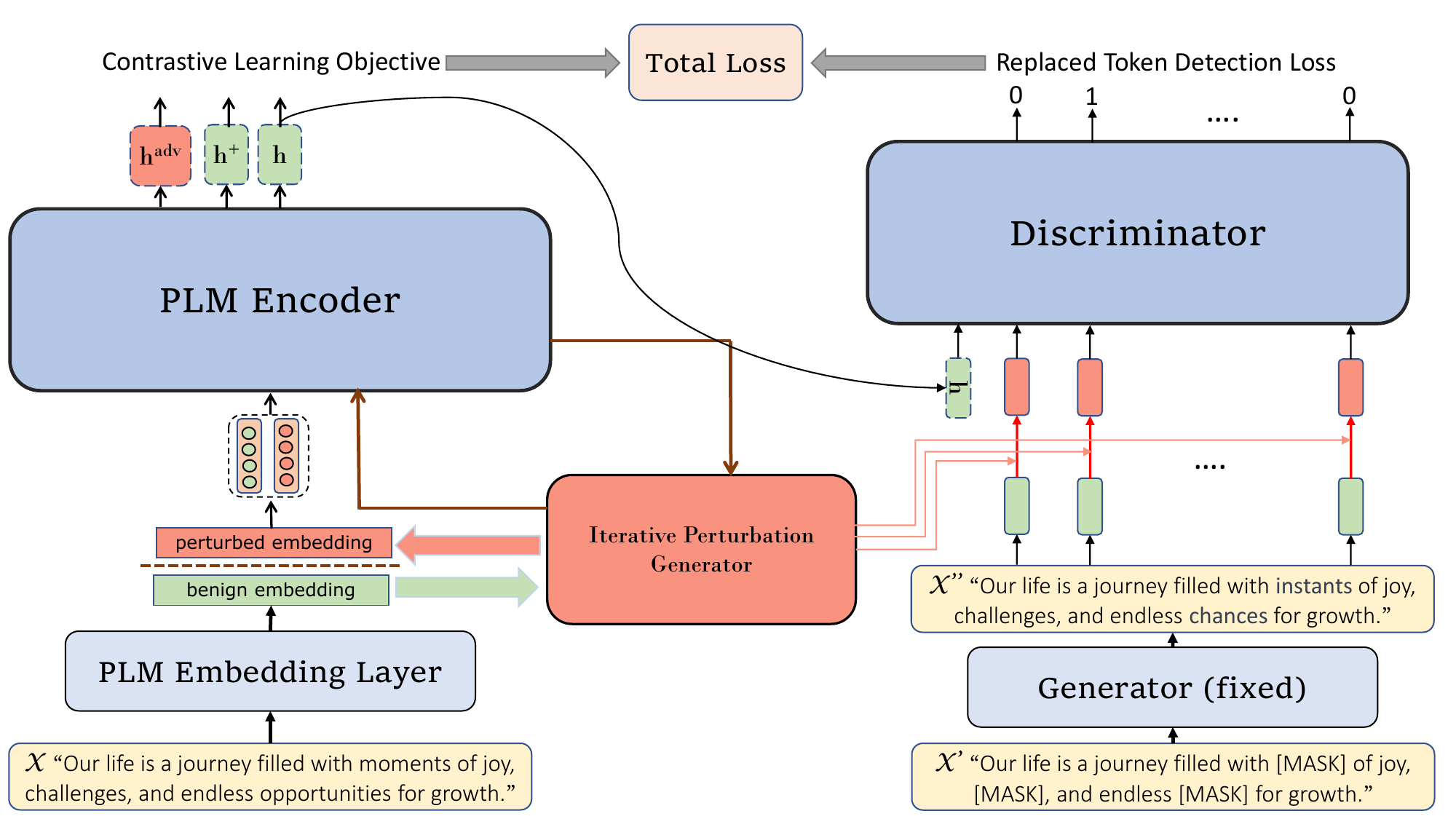}
    \caption{The general architecture of the RobustSentEmbed framework.}
    \label{fig:fig1}
\end{figure*}


\section{The Proposed Framework}
We introduce RobustSentEmbed, a straightforward yet highly effective method for generating robust text representation. Given a PLM $f_\theta(.)$ as the encoder and a raw dataset $\boldsymbol{\mathrm{\mathcal{D}}}$, our framework aims to pre-train $f_\theta(\cdot)$ on $\boldsymbol{\mathrm{\mathcal{D}}}$ to enhance the efficacy of sentence embeddings across a wide range of NLP tasks (improved generalization) and to fortify its resilience against various adversarial attacks (improved robustness). Figure \ref{fig:fig1} presents an overview of our framework. The framework involves an iterative interaction between the perturbation generator and the $f_\theta(.)$ encoder to produce high-risk adversarial perturbations in both token-level and sentence-level embedding spaces. These perturbations provide the essential adversarial examples required for adversarial training by both the $f_\theta(.)$ encoder and a PLM-based discriminator. The subsequent sections will delve into the main components of our framework.


\subsection{Perturbation Generator}
Adversarial perturbation involves adding maliciously crafted perturbations into benign data, with the objective of misleading Machine Learning (ML) models \cite{goodfellow2014explaining}. A highly effective and broadly applicable method for generating adversarial perturbations is to apply a small noise $\boldsymbol{\delta}$ within a norm-constraint ball, aiming to maximize the adversarial loss function:
\begin{equation}
  \arg \max_{{||\boldsymbol{\delta}||\le\epsilon}} L(f_\theta(X+ \boldsymbol{\delta}), y),
\end{equation}
where $f_\theta(.)$ denotes an ML model parameterized with {$X$} as the sub-word embeddings. There are numerous gradient-based algorithms designed to address this optimization problem. Our framework extends the token-level perturbation method proposed by \citet{li2021token} by complementing the perturbation with an innovative sentence-level perturbation generator in order to generate worst-case adversarial examples. The main idea is to train a PLM-based model to withstand a broad spectrum of adversarial attacks, spanning both word and instance levels.



Recognizing the different roles that individual tokens play within a sentence, the RobustSentEmbed framework incorporates a scaling index to allow larger perturbations for tokens exhibiting larger gradients during the normalization of token-level perturbations:
\begin{equation}
    n^i = \frac{\lVert \boldsymbol{\eta}^{t}_{i} \rVert_{P}}{\max_{j} \lVert \boldsymbol{\eta}^{t}_{j} \rVert_{P}},
\end{equation}
where $\boldsymbol{\eta}^{t}_{i}$ represents the token-level perturbation for word $i$ at step $t$ of the gradient ascent, and $P$ denotes the type of norm constraint. Considering the encoder $f_\theta(.)$ and an input sentence $x$, RobustSentEmbed passes the sentence through $f_\theta(.)$ by applying standard dropout twice. This process yields two different embeddings, denoted as "positive pairs" and represented as $({X}, {X^{+}})$. Finally, the newly adjusted token-level perturbation is formulated as:   

\begin{equation}
    \boldsymbol{\eta}^{t+1}_{i}= n^i*(\boldsymbol{\eta}^{t}_{i} + \gamma \frac{\boldsymbol{g}_{{\eta}_{i}}}{\lVert \boldsymbol{g}_{{\eta}_{i}}\rVert_{P})},
\end{equation}            
\begin{equation}
    \boldsymbol{\eta}^{t+1} \gets \Pi_{\lVert \boldsymbol{\eta} \rVert_{P}\leq\epsilon} (\boldsymbol{\eta}^{t}),
\end{equation}
where
\begin{math}
  \boldsymbol{g}_{{\eta}_{i}}= \nabla_{\eta}{\mathcal{L}}_{con,\theta}(\boldsymbol{X} +\boldsymbol{\delta}^{t-1} +\boldsymbol{\eta}^{t-1}, \{\boldsymbol{X}^{+}\})
\end{math} is the gradient of the
contrastive learning loss with respect to $\boldsymbol{\eta}$. The perturbation is generated by the $\ell_{\infty}$ norm-ball with radius $\epsilon$, and $\Pi$ projects the perturbation onto the $\epsilon$-ball.

To generate adversarial perturbations at the sentence-level, RobustSentEmbed employs a combination of the Fast Gradient Sign Method (FGSM) \cite{goodfellow2014explaining} and the Projected Gradient Descent (PGD) technique \cite{madry2017towards}. The framework iterates using this combination, specifically T-step FGSM and K-step PGD, to systematically reinforce invariance within the embedding space. Ultimately, this strategy leads to enhanced generalization and robustness. It proceeds with the following steps to update the perturbation for PGD in iteration $k+1$ and FGSM in iteration $t+1$:
\begin{equation}
\boldsymbol{\delta}^{k+1}_\mathrm{pgd}=\Pi_{\lVert \boldsymbol{\delta} \rVert_{P}\leq\epsilon}(\boldsymbol{\delta}^{k} + \alpha g(\boldsymbol{\delta}^k)/\lVert g(\boldsymbol{\delta}^k) \rVert_{P}), 
\end{equation}
\begin{equation}
\boldsymbol{\delta}^{t+1}_\mathrm{fgsm}=\Pi_{\lVert \boldsymbol{\delta} \rVert_{P}\leq\epsilon}(\boldsymbol{\delta}^{t} + \beta \mathrm{sign}(g(\boldsymbol{\delta}^t))),
\end{equation}
where
\begin{math}
  g(\boldsymbol{\delta}^{n})= \nabla_{\boldsymbol\delta}{\mathcal{L}}_{con,\theta}(\boldsymbol{X}+\boldsymbol{\delta}^{n}, \{\boldsymbol{X}^{+}\})
\end{math} with $n=t\;or\;k$ represents the gradient of the contrastive learning loss with respect to $\boldsymbol{\delta}$. The variables $\alpha$ and $\beta$ denote the step sizes for the attacks, while $\mathrm{sign}(.)$ yields the vector's sign. The final perturbation is obtained by employing a practical combination of T-step FGSM and K-step PGD:
\begin{equation}
  \boldsymbol{\delta}_\mathrm{final}= \rho\boldsymbol{\delta}^{K}_\mathrm{pgd} +(1-\rho)\boldsymbol{\delta}^{T}_\mathrm{fgsm},
  \label{final_pert}
\end{equation}
where $0\leq\rho\leq1$ modulates the relative importance of each separate perturbation in the formation of the final perturbation.

\subsection{Robust Contrastive Learning}
To achieve robust text representations through adversarial learning, we employ a straightforward approach that can be described as the combination of a Replaced Token Detection (RTD) objective (Figure \ref{fig:fig1}, right) with a novel self-supervised contrastive learning objective (Figure \ref{fig:fig1}, left).

Our framework extends an adversarial version of the RTD task used in ELECTRA \cite{clark2020electra}. In this approach, given an input sentence $x$, ELECTRA utilizes a pre-trained masked language model as the generator $G$ to recover randomly masked tokens in $x^{'}=\mathrm{Mask}(x)$, resulting in the edited sentence $x^{''}=G(x^{'})$. Subsequently, a discriminator $D$ is tasked with predicting whether token replacements have occurred, which constitutes the RTD task. As illustrated in Figure \ref{fig:fig1}, the perturbation generator module introduces token-aware perturbations into the embedding of each individual token, making it more challenging for discriminator $D$ to perform the RTD task effectively. The gradient of $D$ can be back-propagated into $f$ through $\boldsymbol{h}=f_\theta(x)$. This mechanism encourages $f$ to make vector $\boldsymbol{h}$ sufficiently informative, enhancing its resilience against token-level adversarial attacks. Consequently, our framework employs the following adversarial objective for a single sentence $x$:

{\small
\begin{flalign}
    {\mathcal{L}}^{x}_{RTD} &=  \sum_{j=1}^{\lvert x \rvert} [\mathds{-1} (X^{adv}_{j} = {X}_j) \; \mathrm{log} \: D (X^{adv},\, \boldsymbol{h},\, j) \nonumber \\
         & \mathds{-1} (X^{adv}_{j} \neq {X}_j) \; \mathrm{log} \: (1-D (X^{adv},\, \boldsymbol{h},\, j))],
\end{flalign}
}
where
\begin{math}
    X^{adv}= {X}^{''} + \boldsymbol{\eta}^{max(K, \:T)}_{i} 
\end{math} represent the $i$th perturbed token in $x$. The training objective for the batch $B$ is 
\begin{math}
    {\mathcal{L}}_{RTD,\,\theta}= \sum_{i=1}^{\lvert B \rvert} {\mathcal{L}}^{x_i}_{RTD} 
\end{math}. 
Furthermore, we use self-supervised contrastive learning to acquire effective low-dimensional representations by bringing semantically similar samples closer and pushing dissimilar ones further apart. Let $\{(x_i, x_i^{+})\}_{i=1}^N$ denote a set of $N$ positive pairs, where $x_i$ and $x_i^{+}$ are semantically correlated and $(z_i, z_i^{+})$ represents the corresponding embedding vectors for the positive pair $(x_i, x_i^{+})$. We define $z_i$'s positive set as $z_i^{pos} = \{z_i^{+}\}$, while the negative set $z_i^{neg}=\{z_i^{-}\}$ is the set of positive pairs from other sentences in the same batch. Then, the contrastive training objective is defined as follows: 

{\small
\begin{flalign}
\label{CLobjective}
& {\mathcal{L}}_{con,\theta}(z_i, z_i^{pos}, z_i^{neg}) = \nonumber \\
& -\log(\frac{\sum_{z_i^{pos}} \exp({sim(z_i, z_i^{+})}/\tau)}{\sum_{(z_i^{pos}\cup\,z_i^{neg})} \exp({sim(z_i, z_i^{+ \, or \, -})}/\tau)}),
\end{flalign}
}

where $\tau$ denotes a temperature hyperparameter and 
\begin{math}
sim(u , v)=\frac{u^{\top}v}{\Arrowvert{u}\Arrowvert.\Arrowvert{v}\Arrowvert} 
\end{math}
is the cosine similarity between two representations. Our framework utilizes contrastive learning to maximize the similarity between clean examples and their adversarial perturbation by incorporating the adversarial example as an additional element within the positive set: 
\begin{equation*}
\label{eqn:CLobjective}
\scalebox{0.89}{$%
{\mathcal{L}}_{RobustSentEmbed,\;\theta} := {\mathcal{L}}_{con,\theta}(z, \{z^{pos},\;z^{adv}\}, \{z^{neg}\}).
$}
\end{equation*}
\begin{equation}
\label{total_loss}
\scalebox{0.76}{$%
\begin{split}
    \mathcal{L}_{total} := & \mathcal{L}_{RobustSentEmbed,\,\theta} + {\lambda_1}\cdot{\mathcal{L}}_{con,\,\theta}(z^{adv}, \{z^{pos}\},\{z^{neg}\}) \\
    & + {\lambda_2}\cdot{\mathcal{L}}_{RTD,\,\theta}, 
\end{split} $}
\end{equation}
where $z^{adv}=z+\boldsymbol{\delta}_{final}$ represents the adversarial perturbation of the input sample $x$ in the embedding space, and {$\lambda_1$, $\lambda_2$} denote weighting coefficients. The first component of the total contrastive loss (Eq. \ref{total_loss}) is designed to optimize the sentence-level similarity between the input sample $x$, its positive pair, and its adversarial perturbation, while the second component serves to regularize the loss by encouraging the convergence of the adversarial perturbation and the positive pair of $x$. The final component introduces the adversarial Replaced Token Detection (RTD) objective into the total contrastive loss.

\begin{table*}[!t]
\renewcommand{\arraystretch}{1.2}
\centering
\Huge
\begin{center}
\resizebox{\linewidth}{!}{%
\begin{tabular}{lccccccccc}
\hline
\bfseries Adversarial Attack& \bfseries Model& \bfseries IMDB & \bfseries MR & \bfseries SST2 & \bfseries YELP & \bfseries MRPC & \bfseries SNLI & \bfseries MNLI-Mismatched & \bfseries Avg.\\
\hline\hline

\multirow{3}{*}{TextFooler} & SimCSE-BERT\textsubscript{base} & 75.32 & 65.53 & 71.49 & 79.67 & 80.07 & 72.65 & 68.54 & 72.61 \\
& USCAL-BERT\textsubscript{base} & 61.94 & 48.71 & 55.38 & 62.30 & 60.18 & 54.82 & 53.74 & 56.72 \\
& RobustSentEmbed-BERT\textsubscript{base} & \textbf{40.02} & \textbf{31.39} & \textbf{35.83} & \textbf{43.78} & \textbf{37.54} & \textbf{36.99} & \textbf{34.15} & \textbf{37.10} \\
\hline

\multirow{3}{*}{TextBugger} & SimCSE-BERT\textsubscript{base} & 52.21   & 42.04 & 49.67 & 56.19 & 56.73  & 45.39 & 40.16 & 48.91 \\
& USCAL-BERT\textsubscript{base}  & 39.16 & 27.37 & 31.90 & 41.25 & 37.86  & 30.79 & 25.45 & 33.40 \\
& RobustSentEmbed-BERT\textsubscript{base} & \textbf{23.16} & \textbf{17.49}  & \textbf{19.62} & \textbf{27.93} & \textbf{19.37} & \textbf{18.05} & \textbf{15.51} & \textbf{20.16} \\
\hline

\multirow{3}{*}{PWWS} & SimCSE-BERT\textsubscript{base}  & 64.41 & 55.73 & 60.48 & 67.54 & 68.15 & 56.09  & 52.58 & 60.71 \\
& USCAL-BERT\textsubscript{base}  & 51.95 & 40.67 & 45.29 & 52.30 & 46.86  & 50.92 & 39.37 & 46.77 \\
& RobustSentEmbed-BERT\textsubscript{base} & \textbf{32.94} & \textbf{28.05}  & \textbf{29.28}  & \textbf{29.14} & \textbf{24.72} & \textbf{26.28}  & \textbf{27.90} & \textbf{28.33} \\
\hline

\multirow{3}{*}{BAE} & SimCSE-BERT\textsubscript{base} & 73.50 & 61.83 & 68.27 & 75.15 & 77.84 & 69.06 & 65.43 & 70.15 \\
& USCAL-BERT\textsubscript{base} & 58.57 & 46.19 & 51.72 & 59.49 & 58.38 & 50.90 & 51.16 & 53.77 \\
& RobustSentEmbed-BERT\textsubscript{base} & \textbf{37.16} & \textbf{29.12} & \textbf{31.43} & \textbf{40.96} & \textbf{35.53} & \textbf{33.87} & \textbf{31.85} & \textbf{34.27} \\
\hline

\multirow{3}{*}{BERTAttack} & SimCSE-BERT\textsubscript{base} & 78.42 & 66.94 & 73.59 & 80.87 & 82.16 & 74.35 & 72.22 & 75.51 \\
& USCAL-BERT\textsubscript{base} & 63.23 & 51.08 & 57.73 & 63.96 & 63.05 & 55.41 & 55.86 & 58.62 \\
& RobustSentEmbed-BERT\textsubscript{base} & \textbf{41.51} & \textbf{34.19} & \textbf{38.16} & \textbf{44.96} & \textbf{38.26} & \textbf{38.60} & \textbf{35.98} & \textbf{38.81} \\
\hline

\end{tabular}
}
\end{center}
\caption{Attack success rates (lower is better) of various adversarial attacks applied to three sentence embeddings (SimCSE, USCAL, and RobustSentEmbed) across five text classification and two natural language inference tasks. RobustSentEmbed reduces the attack success rate to less than half across all attacks.}
\label{tbl3}
\end{table*}

\section{Evaluation and Experimental Results}
This section presents a comprehensive set of experiments conducted to validate the proposed framework's effectiveness in terms of robustness and generalization metrics. To evaluate robustness, the experiments include adversarial attacks and adversarial Semantic Textual Similarity (STS) tasks. To evaluate generalization, the experiments include non-adversarial STS and transfer tasks within the SentEval framework.\footnote{\url{https://github.com/facebookresearch/SentEval}} Appendices \ref{Training} and \ref{Ablation} provide training details and ablation studies that illustrate the effects of hyperparameter tuning.


\begin{table*}[!t]
\centering
\tiny
\begin{center}
\resizebox{\linewidth}{!}{%
\begin{tabular}{lccccccccc}
\hline
\bfseries Adversarial Attack& \bfseries Model& \bfseries AdvSTS-B & \bfseries AdvSICK-R & \bfseries Avg.\\
\hline\hline

\multirow{3}{*}{TextFooler} & SimCSE-BERT\textsubscript{base} & 21.07 & 24.17 & 22.62 \\
& USCAL-BERT\textsubscript{base} & 16.52 & 18.71 & 17.62 \\
& RobustSentEmbed-BERT\textsubscript{base} & \textbf{7.18} & \textbf{8.53} & \textbf{7.86}  \\
\hline

\multirow{3}{*}{TextBugger} & SimCSE-BERT\textsubscript{base} & 27.49 & 28.34 & 27.91 \\
& USCAL-BERT\textsubscript{base} & 21.52 & 24.88 & 23.20 \\
& RobustSentEmbed-BERT\textsubscript{base} & \textbf{11.32} & \textbf{12.94} & \textbf{12.13}  \\
\hline

\multirow{3}{*}{PWWS} & SimCSE-BERT\textsubscript{base} & 24.15 & 26.82 & 25.49 \\
& USCAL-BERT\textsubscript{base} & 21.28 & 23.65 & 22.47 \\
& RobustSentEmbed-BERT\textsubscript{base} & \textbf{12.68} & \textbf{13.90} & \textbf{13.29}  \\
\hline

\multirow{3}{*}{BAE} & SimCSE-BERT\textsubscript{base} & 26.92 & 28.81 & 27.86 \\
& USCAL-BERT\textsubscript{base} & 22.92 & 25.48 & 24.20 \\
& RobustSentEmbed-BERT\textsubscript{base} & \textbf{10.53} & \textbf{12.09} & \textbf{11.31}  \\
\hline

\multirow{3}{*}{BERTAttack} & SimCSE-BERT\textsubscript{base} & 31.60 & 32.85 & 32.23 \\
& USCAL-BERT\textsubscript{base} & 26.02 & 28.51 & 27.26 \\
& RobustSentEmbed-BERT\textsubscript{base} & \textbf{12.58} & \textbf{13.02} & \textbf{12.80}  \\
\hline

\end{tabular}
}
\end{center}
\caption{Attack success rates (lower is better) of five adversarial attack techniques applied to three sentence embeddings (SimCSE, USCAL, and RobustSentEmbed) across two Adversarial Semantic Textual Similarity (AdvSTS) tasks (i.e. AdvSTS-B and AdvSICK-R). RobustSentEmbed reduces the attack success rate to less than half across all attacks.}
\label{tbl4}
\end{table*}

\subsection{Adversarial Attacks}
\label{Attack_ex}
We evaluate the robustness of our framework against various adversarial attacks, comparing it with two state-of-the-art sentence embedding models: SimSCE \cite{gao2021simcse} and USCAL \cite{miao2021simple}. We fine-tuned the BERT-based PLM across seven text classification and natural language inference tasks, specifically MRPC \cite{dolan_brockett_2005_automatically}, YELP \cite{zhang2015character}, IMDb \cite{IMDb}, Movie Reviews (MR) \cite{pang2005seeing}, SST2 \cite{socher_etal_2013_recursive}, Stanford NLI (SNLI) \cite{bowman2015large}, and Multi-NLI (MNLI) \cite{williams2017broad}. To assess the robustness of our fine-tuned model, we investigated the impact of five popular adversarial attacks: TextBugger \cite{li2018textbugger}, PWWS \cite{ren_etal_2019_generating}, TextFooler \cite{jin2020bert}, BAE \cite{garg2020bae}, and BERTAttack \cite{li_etal_2020_bert_attack}. Additional information of these attacks is provided in Appendix \ref{attacks}. To ensure statistical validity, we conducted each experiment five times, with each iteration comprising 1000 adversarial attack samples.


Table \ref{tbl3} presents the average attack success rates of five adversarial attacks applied to three sentence embeddings. Notably, our embedding framework consistently outperforms the other two embedding methods, demonstrating significantly lower attack success rates (less than half) across all text classification and natural language inference tasks. Consequently, RobustSentEmbed achieves the lowest average attack success rate against all adversarial attack techniques. These findings substantiate the robustness of our embedding framework and highlight the vulnerabilities of other state-of-the-art sentence embeddings when confronted with various adversarial attacks.

\begin{figure}[t]
    \centering
    \includegraphics[width=\columnwidth, height=6cm]{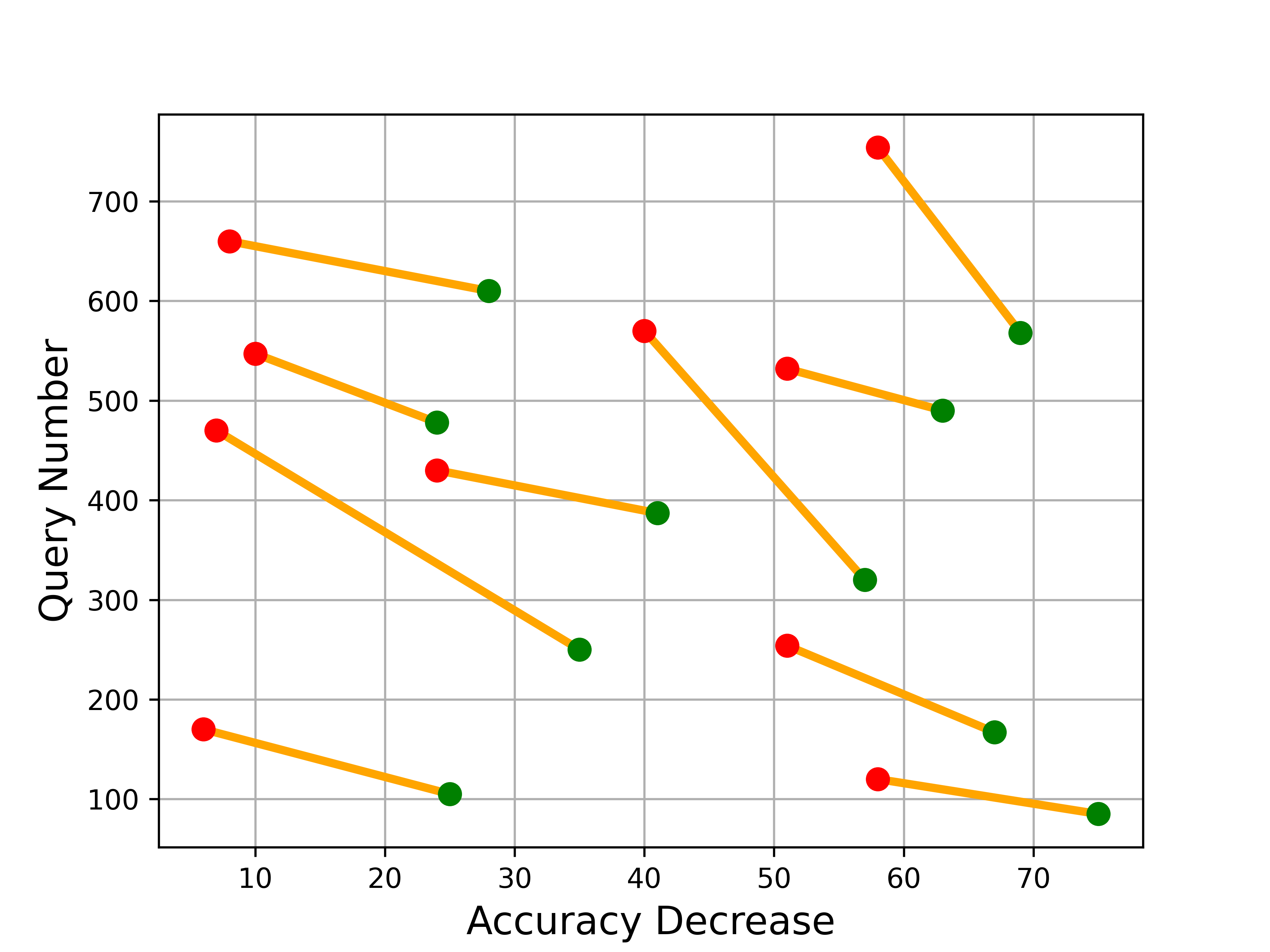}
    \caption{Average number of queries and the resulting accuracy reduction for two fine-tuned embeddings.}
    \label{fig:fig2}
\end{figure}

Figure \ref{fig:fig2} presents the results of 1000 attacks conducted on two fine-tuned sentence embeddings, assessing the average number of queries required and the resulting accuracy reduction. Attacks on the RobustSentEmbed framework are represented by green data points, while red points denote attacks on the USCAL approach \cite{miao2021simple}. Each pair of connected points corresponds to a specific attack. Ideally, a robust sentence embedding should be positioned in the top-left region of the graph, indicating that it necessitates a higher number of queries for an attack to deceive the model while causing minimal performance degradation. Across all adversarial attacks, RobustSentEmbed consistently exhibits greater stability compared to the USCAL method. In other words, a larger number of queries is required for RobustSentEmbed, resulting in a lower accuracy reduction (i.e., better performance) compared to USCAL. 


\subsection{Robust Embeddings}
\label{Robust_ex}
We introduce a new task named Adversarial Semantic Textual Similarity (AdvSTS) to assess the robustness of sentence embeddings. AdvSTS leverages an efficient adversarial technique, like TextFooler, to manipulate an input sentence pair of a non-adversarial STS task in a manner that leads the target model to generate a regression score that maximally deviates from the actual score (truth label). As a result, we generate an adversarial STS dataset by transforming all benign instances from the original (i.e. non-adversarial) dataset into adversarial examples. Table \ref{tbl4} presents the attack success rates of five adversarial attacks applied to three sentence embeddings, including our framework. These evaluations are conducted for two AdvSTS tasks, specifically AdvSTS-B (originated from STS Benchmark \cite{cer-etal-2017-semeval}) and AdvSICK-R (originated from SICK-Relatedness \cite{ marelli2014sick}). Notably, our framework consistently outperforms the other two sentence embedding methods, exhibiting significantly lower attack success rates across both AdvSTS tasks and all employed adversarial attacks. These results provide additional evidence supporting the notion that RobustSentEmbed generates robust text representation.


\begin{table*}
  \resizebox{\linewidth}{!}{%
  \begin{tabular}{lcccccccc}
\hline
    \bfseries Model & \bfseries STS12 & \bfseries STS13 & \bfseries STS14 & \bfseries STS15 & \bfseries STS16 & \bfseries STS-B & \bfseries  SICK-R & \bfseries Avg.\\
\hline
\hline   
    
    \texorpdfstring{GloVe embeddings (avg.)	\textsuperscript{$\heartsuit$}} && 55.14 & 70.66 & 59.73 &  68.25 &  63.66 & 58.02 & 53.76 & 61.32 \\

    \texorpdfstring{BERT\textsubscript{base} (first-last avg.) \textsuperscript{$\clubsuit$}} &&  39.70 & 59.38 & 49.67 & 66.03 & 66.19 & 53.87 & 62.06 & 56.70 \\

    \texorpdfstring{BERT\textsubscript{base}-flow \textsuperscript{$\clubsuit$}} && 58.40 & 67.10 & 60.85 & 75.16 & 71.22 & 68.66 & 64.47 & 66.55 \\
    
    \texorpdfstring{BERT\textsubscript{base}-whitening \textsuperscript{$\clubsuit$}} && 57.83 & 66.90 & 60.90 & 75.08 & 71.31 & 68.24 & 63.73 & 66.28 \\

    \texorpdfstring{ConSERT-BERT\textsubscript{base}} && 64.56 & 78.55 & 69.16 & 79.74 & 76.00 & 73.91 & 67.35 & 72.75 \\
    
    \texorpdfstring{ATCL-BERT\textsubscript{base}} && 67.14 & 80.86 &  71.73 & 79.50 & 76.72 & 79.31 & 70.49 & 75.11 \\  
    
    \texorpdfstring{SimCSE-BERT\textsubscript{base}} && 68.66 & \textbf{81.73} & 72.04 & 80.53 & 78.09 & 79.94 & 71.42 & 76.06 \\  
    
    \texorpdfstring{USCAL-BERT\textsubscript{base}} && 69.30 & 80.85 & 72.19 & 81.04 & 77.52 & 81.28 & 71.98 & 76.31 \\
        
    \texorpdfstring{RobustSentEmbed-BERT\textsubscript{base}} && \textbf{71.90} & 81.12 & \textbf{74.92} & \textbf{82.38} & \textbf{79.43} & \textbf{82.02} & \textbf{73.53} & \textbf{77.90} \\  
    
\hline
    \texorpdfstring{RoBERTa\textsubscript{base}-whitening} && 46.99 & 63.24 &  57.23 & 71.36 &  68.99 & 61.36 & 62.91 & 61.73 \\

    \texorpdfstring{ConSERT-RoBERTa\textsubscript{base}} && 66.90 & 79.31 & 70.33 & 80.57 & 77.95 & 81.42 & 68.16 & 74.95 \\

    \texorpdfstring{SimCSE-RoBERTa\textsubscript{base}} && 68.75 & 80.81 & 71.19 & 81.79 & 79.35 & 82.62 & 69.56 & 76.30 \\
 
     \texorpdfstring{USCAL-RoBERTa\textsubscript{base}} && 69.28 & 81.15 & 72.81 & 81.47 & \textbf{80.55} & 83.34 & 70.94 & 77.08 \\  
 
      \texorpdfstring{RobustSentEmbed-RoBERTa\textsubscript{base}} && \textbf{70.03} & \textbf{82.15} & \textbf{73.27} & \textbf{82.48} & 79.61 & \textbf{83.82} & \textbf{71.66} & \textbf{77.57} \\  
      
\hline
    \texorpdfstring{USCAL-RoBERTa\textsubscript{large}} && 68.70 & \textbf{81.84} & 74.26 & 82.52 & \textbf{80.01} & 83.14 & 76.30 & 78.11 \\  
 
    \texorpdfstring{RobustSentEmbed-RoBERTa\textsubscript{large}} && \textbf{69.30} & 81.76 & \textbf{75.14} & \textbf{83.57} & 79.74 & \textbf{83.90} & \textbf{77.08} & \textbf{78.64} \\  
     
\hline
  \end{tabular}
  }
\caption{Semantic Similarity performance on STS tasks (Spearman’s correlation, “all” setting) for sentence embedding models. We emphasize the top-performing numbers among models that share the same pre-trained encoder. $\heartsuit$: results from \citet{reimers2019sentence}; $\clubsuit$: results from \cite{gao2021simcse}; All remaining results have been reproduced and reevaluated by our team. RobustSentEmbed produces the most effective sentence representations that are more general in addition to robust representation (section \ref{Robust_ex} and \ref{Attack_ex}).}
\label{tab:tbl1}
\end{table*}

\subsection{Semantic Textual Similarity (STS) Tasks}
In this section, we assess the performance of our framework across seven Semantic Textual Similarity (STS) tasks encompassing STS datasets from 2012 to 2016 \cite{agirre-etal-2012-semeval, agirre-etal-2013-sem, agirre-etal-2014-semeval,agirre2015semeval,agirre-etal-2016-semeval}, STS Benchmark, and SICK-Relatedness. To benchmark our framework's effectiveness, we conducted a comparative analysis against a range of unsupervised sentence embedding approaches, including: 1) baseline methods such as GloVe \cite{pennington_etal_2014_glove} and average BERT embeddings; 2) post-processing methods like BERT-flow \cite{li2020sentence} and BERT-whitening \cite{su2021whitening}; and 3) state-of-the-art methods such as SimCSE \cite{gao2021simcse} and USCAL \cite{miao2021simple}. We validate the findings of the SimCSE, ConSERT, and USCAL frameworks by replicating their results. The empirical outcomes, as presented in Table \ref{tab:tbl1}, consistently establish the superior performance of our RobustSentEmbed framework in contrast to various other sentence embeddings. Our framework achieves the highest average Spearman's correlation score when compared to state-of-the-art approaches. Specifically, utilizing the BERT encoder, our framework surpasses the second-best embedding method, USCAL, by a margin of 1.59\%. Moreover, RobustSentEmbed achieves the highest score in the majority of individual STS tasks, outperforming other embedding methods in 6 out of 7 tasks. For the RoBERTa encoder, RobustSentEmbed outperforms the state-of-the-art embeddings in five out of seven STS tasks and attains the highest average Spearman's correlation score.


\begin{table*}
  \resizebox{\linewidth}{!}{%
  
  \begin{tabular}{lccccccc}
\hline
    \bfseries Model & \bfseries MR & \bfseries CR & \bfseries SUBJ & \bfseries MPQA & \bfseries SST2 & \bfseries MRPC & \bfseries Avg.\\
\hline
\hline
    
     \texorpdfstring{GloVe embeddings (avg.)	\textsuperscript{$\clubsuit$}} && 77.25 &  78.30 & 91.17 & 87.85 & 80.18 & 72.87 & 81.27 \\

     \texorpdfstring{Skip-thought \textsuperscript{$\heartsuit$}} && 76.50 &  80.10 & 93.60 & 87.10 & 82.00 & 73.00 & 82.05 \\
     
\hline
  
     \texorpdfstring{BERT-{\fontfamily{qcr}\selectfont[CLS]} embedding	\textsuperscript{$\clubsuit$}} && 78.68 &  84.85 & 94.21 & 88.23 & 84.13 & 71.13 & 83.54 \\

     \texorpdfstring{ConSERT-BERT\textsubscript{base}} && 79.52 &  87.05 & 94.32 & 88.47 & 85.46 & 72.54 & 84.56 \\
     
    \texorpdfstring{SimCSE-BERT\textsubscript{base}} && 81.29 & 86.94 & 94.72 & 89.49 & \textbf{86.70} & 75.13 & 85.71 \\ 
    
     \texorpdfstring{USCAL-BERT\textsubscript{base}} && 81.54 & \textbf{87.12} & 95.24 & 89.34 & 85.71 & 75.84 & 85.80 \\ 
    	  
    \texorpdfstring{RobustSentEmbed-BERT\textsubscript{base}} && \textbf{82.06} & 86.28 & \textbf{95.42} & \textbf{89.61} & 86.12 & \textbf{76.69} & \textbf{86.03} \\ 
    
\hline
    
    \texorpdfstring{SimCSE-RoBERTa\textsubscript{base}} && 81.15 & 87.15 & 92.38 & 86.79 & \textbf{86.24} & 75.49 & 84.87 \\  

    \texorpdfstring{USCAL-RoBERTa\textsubscript{base}} && \textbf{82.15} & 87.22 & 92.76 & 87.74 & 84.39 & 76.20 & 85.08 \\  
        
    \texorpdfstring{RobustSentEmbed-RoBERTa\textsubscript{base}} && 81.57 & \textbf{87.66} & \textbf{93.51} & \textbf{87.94} & 85.04 & \textbf{76.89} & \textbf{85.44} \\ 
    
\hline
    
    \texorpdfstring{USCAL-RoBERTa\textsubscript{large}} && \textbf{82.84} & 87.97 & 93.12 & 88.48 & \textbf{86.28} & 76.41 & 85.85 \\  
    	  
    \texorpdfstring{RobustSentEmbed-RoBERTa\textsubscript{large}} && 82.56 & \textbf{88.51} & \textbf{93.84} & \textbf{88.65} & 86.18 & \textbf{77.01} & \textbf{86.13} \\ 
    
\hline
  \end{tabular}
  }
\caption{Results of transfer tasks for different sentence embedding models. $\clubsuit$:
results from \citet{reimers2019sentence}; $\heartsuit$: results from \citet{zhang2020unsupervised}; We emphasize the top-performing numbers among models that share the same pre-trained encoder. All remaining results have been reproduced and reevaluated by our team. RobustSentEmbed outperforms all other methods, regardless of the pre-trained language model (BERT\textsubscript{base}, RoBERTa\textsubscript{base}, or RoBERTa\textsubscript{large}).}
\label{tab:tbl2}
\end{table*}

\subsection{Transfer Tasks}
We leveraged transfer tasks to assess the performance of our framework, RobustSentEmbed, across a diverse range of text classification tasks, including sentiment analysis and paraphrase identification. Our evaluation encompassed six transfer tasks: CR \cite{hu2004mining}, SUBJ \cite{pang2004sentimental}, MPQA \cite{wiebe2005annotating}, SST2 \cite{socher_etal_2013_recursive}, and MRPC \cite{dolan_brockett_2005_automatically}. We trained a logistic regression classifier on top of the fixed sentence embeddings. To ensure the reliability of our findings, we replicated the SimCSE, ConSERT, and USCAL frameworks. The outcomes, as presented in Table \ref{tab:tbl2}, demonstrate the superior performance of our framework in terms of average accuracy when compared to other sentence embeddings. Specifically, when utilizing the BERT encoder, our framework outperforms the second-best embedding method by a margin of 0.23\%. Furthermore, RobustSentEmbed achieves the highest score in four out of six text classification tasks. A similar trend is observed for the RoBERTa encoder. Overall, based on the results presented in Tables \ref{tab:tbl1} and \ref{tab:tbl2}, we conclude that RobustSentEmbed generates general sentence representation in addition to robust representation (\ref{Attack_ex} and section \ref{Robust_ex} ). 


In conclusion, the comprehensive experiments, as indicated by the outcomes in Tables \ref{tbl3}, \ref{tbl4}, \ref{tab:tbl1}, and \ref{tab:tbl2}, along with Figure \ref{fig:fig2}, confirm the exceptional performance of RobustSentEmbed in text representation and resilience against adversarial attacks and adversarial tasks. These findings highlight the framework's outstanding robustness and generalization capabilities, underscoring its potential as a versatile method for generating high-quality sentence embeddings.

\subsection{Distribution of Sentence Embeddings}
\begin{figure}[t]
    \centering
    \includegraphics[width=\columnwidth, height=5.5cm]{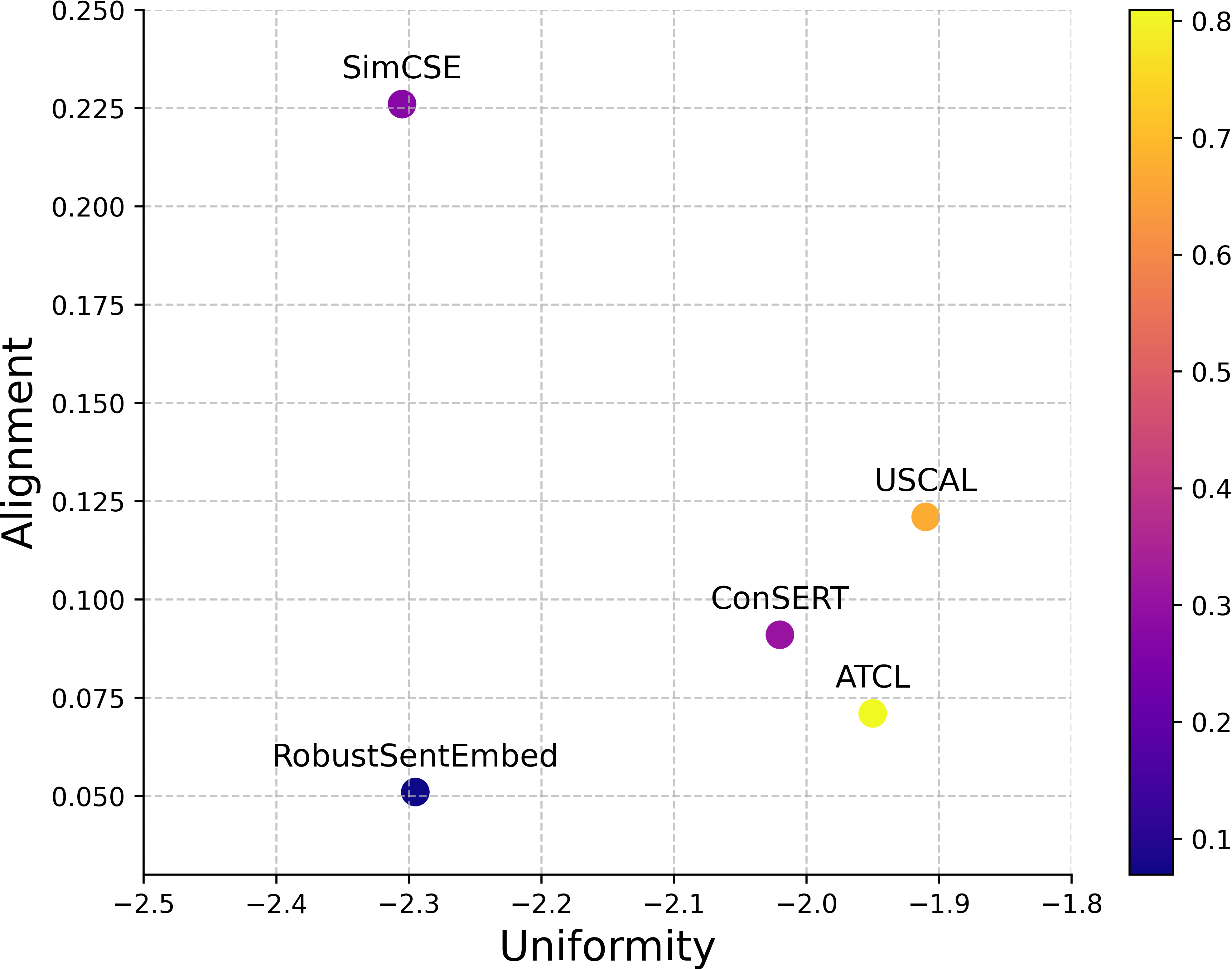}
    \caption{ $\ell_\text{align}-\ell_\text{uniform}$ plot of models based on \texorpdfstring{BERT\textsubscript{base}} e. Lower uniformity and alignment is better.}
    \label{fig:fig3}
\end{figure}

We employed two critical metrics, \textit{alignment} and \textit{uniformity} \cite{wang2020understanding}, for evaluating the quality of our representations. With a distribution of positive pairs $p_{pos}$, \textit{alignment} computes the expected distance between the embeddings of paired instances:
\begin{equation}
 \ell_\text{align}  \triangleq \mathop{\mathbb{E}}_{{(x,x^+)} \thicksim p_{pos}}  \Vert f(x) - f(x^+) \Vert^2
\end{equation}
\textit{Uniformity} measures how well the embeddings are uniformly distributed in the representation space:
\begin{equation}
 \ell_\text{uniform}  \triangleq \log \mathop{\mathbb{E}}_{{x, y  \stackrel{i.i.d.}\thicksim p_{\text{data}}}} e^{-2 \Vert f(x) - f(y) \Vert^2}
\end{equation}
Figure \ref{fig:fig3} shows the \textit{uniformity} and \textit{alignment} of different sentence embedding models. Smaller values indicate better performance. In comparison to the other representations, RobustSentEmbed achieves a similar level of \textit{uniformity} (-2.295 vs. -2.305) but exhibits superior \textit{alignment} (0.051 vs. 0.073). This demonstrates that our framework is more efficient in optimizing the representation space in two different directions.

\section{Conclusion and Future Work}
This paper introduces RobustSentEmbed, a self-supervised sentence embedding framework enhancing robustness against adversarial attacks while achieving state-of-the-art performance in text representation and NLP tasks. Current sentence embeddings are vulnerable to attacks, and RobustSentEmbed addresses this by generating high-risk perturbations at token and sentence levels. These perturbations are incorporated into novel contrastive and difference prediction objectives. The framework is validated through comprehensive experiments on semantic textual similarity and transfer learning tasks, confirming its robustness against adversarial attacks and semantic similarity tasks. In future research, we aim to investigate the use of hard negative examples to  further enhance the effectiveness of text representations.

\section{Limitations}
Despite the effectiveness of our approach and its notable performance, there are potential limitations to our framework:

\begin{itemize}
  \item The framework is primarily tailored for descriptive models like BERT, adept at language understanding and representation, including tasks such as text classification. However, its direct application to generative models like GPT, focused on generating coherent and contextually relevant text, may pose challenges. Thus, applying our methodology to enhance generalization and robustness in generative pre-trained models might have limitations.
  \item Utilizing substantial GPU resources is necessary for pre-training large-scale models like RoBERTa\textsubscript{large} in our framework. Due to limited GPU availability, we had to use smaller batch sizes during pre-training. Although larger batch sizes typically result in better performance, our experiments had to compromise and use smaller batch sizes to efficiently generate sentence embeddings within GPU constraints.
\end{itemize}



\bibliography{acl_latex}

\begin{thebibliography}{53}
\expandafter\ifx\csname natexlab\endcsname\relax\def\natexlab#1{#1}\fi

\bibitem[{Agirre et~al.(2015)Agirre, Banea, Cardie, Cer, Diab, Gonzalez-Agirre, Guo, Lopez-Gazpio, Maritxalar, Mihalcea et~al.}]{agirre2015semeval}
Eneko Agirre, Carmen Banea, Claire Cardie, Daniel Cer, Mona Diab, Aitor Gonzalez-Agirre, Weiwei Guo, Inigo Lopez-Gazpio, Montse Maritxalar, Rada Mihalcea, et~al. 2015.
\newblock Semeval-2015 task 2: Semantic textual similarity, english, spanish and pilot on interpretability.
\newblock In \emph{Proceedings of the 9th international workshop on semantic evaluation (SemEval 2015)}, pages 252--263.

\bibitem[{Agirre et~al.(2014)Agirre, Banea, Cardie, Cer, Diab, Gonzalez-Agirre, Guo, Mihalcea, Rigau, and Wiebe}]{agirre-etal-2014-semeval}
Eneko Agirre, Carmen Banea, Claire Cardie, Daniel Cer, Mona Diab, Aitor Gonzalez-Agirre, Weiwei Guo, Rada Mihalcea, German Rigau, and Janyce Wiebe. 2014.
\newblock \href {https://doi.org/10.3115/v1/S14-2010} {{S}em{E}val-2014 task 10: Multilingual semantic textual similarity}.
\newblock In \emph{Proceedings of the 8th International Workshop on Semantic Evaluation ({S}em{E}val 2014)}, pages 81--91, Dublin, Ireland. Association for Computational Linguistics.

\bibitem[{Agirre et~al.(2016)Agirre, Banea, Cer, Diab, Gonzalez-Agirre, Mihalcea, Rigau, and Wiebe}]{agirre-etal-2016-semeval}
Eneko Agirre, Carmen Banea, Daniel Cer, Mona Diab, Aitor Gonzalez-Agirre, Rada Mihalcea, German Rigau, and Janyce Wiebe. 2016.
\newblock \href {https://doi.org/10.18653/v1/S16-1081} {{S}em{E}val-2016 task 1: Semantic textual similarity, monolingual and cross-lingual evaluation}.
\newblock In \emph{Proceedings of the 10th International Workshop on Semantic Evaluation ({S}em{E}val-2016)}, pages 497--511, San Diego, California. Association for Computational Linguistics.

\bibitem[{Agirre et~al.(2012)Agirre, Cer, Diab, and Gonzalez-Agirre}]{agirre-etal-2012-semeval}
Eneko Agirre, Daniel Cer, Mona Diab, and Aitor Gonzalez-Agirre. 2012.
\newblock \href {https://aclanthology.org/S12-1051} {{S}em{E}val-2012 task 6: A pilot on semantic textual similarity}.
\newblock In \emph{*{SEM} 2012: The First Joint Conference on Lexical and Computational Semantics {--} Volume 1: Proceedings of the main conference and the shared task, and Volume 2: Proceedings of the Sixth International Workshop on Semantic Evaluation ({S}em{E}val 2012)}, pages 385--393, Montr{\'e}al, Canada. Association for Computational Linguistics.

\bibitem[{Agirre et~al.(2013)Agirre, Cer, Diab, Gonzalez-Agirre, and Guo}]{agirre-etal-2013-sem}
Eneko Agirre, Daniel Cer, Mona Diab, Aitor Gonzalez-Agirre, and Weiwei Guo. 2013.
\newblock \href {https://aclanthology.org/S13-1004} {*{SEM} 2013 shared task: Semantic textual similarity}.
\newblock In \emph{Second Joint Conference on Lexical and Computational Semantics (*{SEM}), Volume 1: Proceedings of the Main Conference and the Shared Task: Semantic Textual Similarity}, pages 32--43, Atlanta, Georgia, USA. Association for Computational Linguistics.

\bibitem[{Bowman et~al.(2015)Bowman, Angeli, Potts, and Manning}]{bowman2015large}
Samuel~R. Bowman, Gabor Angeli, Christopher Potts, and Christopher~D. Manning. 2015.
\newblock \href {https://doi.org/10.18653/v1/D15-1075} {A large annotated corpus for learning natural language inference}.
\newblock In \emph{Proceedings of the 2015 Conference on Empirical Methods in Natural Language Processing}, pages 632--642, Lisbon, Portugal. Association for Computational Linguistics.

\bibitem[{Brown et~al.(2020)Brown, Mann, Ryder, Subbiah, Kaplan, Dhariwal, Neelakantan, Shyam, Sastry, Askell et~al.}]{brown2020language}
Tom Brown, Benjamin Mann, Nick Ryder, Melanie Subbiah, Jared~D Kaplan, Prafulla Dhariwal, Arvind Neelakantan, Pranav Shyam, Girish Sastry, Amanda Askell, et~al. 2020.
\newblock Language models are few-shot learners.
\newblock \emph{Advances in neural information processing systems}, 33:1877--1901.

\bibitem[{Cer et~al.(2017)Cer, Diab, Agirre, Lopez-Gazpio, and Specia}]{cer-etal-2017-semeval}
Daniel Cer, Mona Diab, Eneko Agirre, I{\~n}igo Lopez-Gazpio, and Lucia Specia. 2017.
\newblock \href {https://doi.org/10.18653/v1/S17-2001} {{S}em{E}val-2017 task 1: Semantic textual similarity multilingual and crosslingual focused evaluation}.
\newblock In \emph{Proceedings of the 11th International Workshop on Semantic Evaluation ({S}em{E}val-2017)}, pages 1--14, Vancouver, Canada. Association for Computational Linguistics.

\bibitem[{Chen et~al.(2020)Chen, Kornblith, Norouzi, and Hinton}]{chen2020simple}
Ting Chen, Simon Kornblith, Mohammad Norouzi, and Geoffrey Hinton. 2020.
\newblock A simple framework for contrastive learning of visual representations.
\newblock In \emph{International conference on machine learning}, pages 1597--1607. PMLR.

\bibitem[{Clark et~al.(2020)Clark, Luong, Le, and Manning}]{clark2020electra}
Kevin Clark, Minh-Thang Luong, Quoc~V. Le, and Christopher~D. Manning. 2020.
\newblock \href {https://openreview.net/pdf?id=r1xMH1BtvB} {{ELECTRA}: Pre-training text encoders as discriminators rather than generators}.
\newblock In \emph{ICLR}.

\bibitem[{Conneau and Kiela(2018)}]{conneau2018senteval}
Alexis Conneau and Douwe Kiela. 2018.
\newblock \href {https://aclanthology.org/L18-1269} {{S}ent{E}val: An evaluation toolkit for universal sentence representations}.
\newblock In \emph{Proceedings of the Eleventh International Conference on Language Resources and Evaluation ({LREC} 2018)}, Miyazaki, Japan. European Language Resources Association (ELRA).

\bibitem[{Devlin et~al.(2019)Devlin, Chang, Lee, and Toutanova}]{devlin_etal_2019_ber}
Jacob Devlin, Ming-Wei Chang, Kenton Lee, and Kristina Toutanova. 2019.
\newblock \href {https://doi.org/10.18653/v1/N19-1423} {{BERT}: Pre-training of deep bidirectional transformers for language understanding}.
\newblock In \emph{Proceedings of the 2019 Conference of the North {A}merican Chapter of the Association for Computational Linguistics: Human Language Technologies, Volume 1 (Long and Short Papers)}, pages 4171--4186, Minneapolis, Minnesota. Association for Computational Linguistics.

\bibitem[{Ding et~al.(2023)Ding, Qin, Yang, Wei, Yang, Su, Hu, Chen, Chan, Chen et~al.}]{ding2023parameter}
Ning Ding, Yujia Qin, Guang Yang, Fuchao Wei, Zonghan Yang, Yusheng Su, Shengding Hu, Yulin Chen, Chi-Min Chan, Weize Chen, et~al. 2023.
\newblock Parameter-efficient fine-tuning of large-scale pre-trained language models.
\newblock \emph{Nature Machine Intelligence}, 5(3):220--235.

\bibitem[{Dolan and Brockett(2005)}]{dolan_brockett_2005_automatically}
William~B. Dolan and Chris Brockett. 2005.
\newblock \href {https://aclanthology.org/I05-5002} {Automatically constructing a corpus of sentential paraphrases}.
\newblock In \emph{Proceedings of the Third International Workshop on Paraphrasing ({IWP}2005)}.

\bibitem[{Gao et~al.(2021)Gao, Yao, and Chen}]{gao2021simcse}
Tianyu Gao, Xingcheng Yao, and Danqi Chen. 2021.
\newblock \href {https://doi.org/10.18653/v1/2021.emnlp-main.552} {{S}im{CSE}: Simple contrastive learning of sentence embeddings}.
\newblock In \emph{Proceedings of the 2021 Conference on Empirical Methods in Natural Language Processing}, pages 6894--6910, Online and Punta Cana, Dominican Republic. Association for Computational Linguistics.

\bibitem[{Garg and Ramakrishnan(2020)}]{garg2020bae}
Siddhant Garg and Goutham Ramakrishnan. 2020.
\newblock \href {https://doi.org/10.18653/v1/2020.emnlp-main.498} {{BAE}: {BERT}-based adversarial examples for text classification}.
\newblock In \emph{Proceedings of the 2020 Conference on Empirical Methods in Natural Language Processing (EMNLP)}, pages 6174--6181, Online. Association for Computational Linguistics.

\bibitem[{Goodfellow et~al.(2015)Goodfellow, Shlens, and Szegedy}]{goodfellow2014explaining}
Ian~J. Goodfellow, Jonathon Shlens, and Christian Szegedy. 2015.
\newblock \href {http://arxiv.org/abs/1412.6572} {Explaining and harnessing adversarial examples}.
\newblock In \emph{3rd International Conference on Learning Representations, {ICLR} 2015, San Diego, CA, USA, May 7-9, 2015, Conference Track Proceedings}.

\bibitem[{Hauser et~al.(2023)Hauser, Meng, Pascual, and Wattenhofer}]{hauser2023bert}
Jens Hauser, Zhao Meng, Damian Pascual, and Roger Wattenhofer. 2023.
\newblock Bert is robust! a case against word substitution-based adversarial attacks.
\newblock In \emph{ICASSP 2023-2023 IEEE International Conference on Acoustics, Speech and Signal Processing (ICASSP)}, pages 1--5. IEEE.

\bibitem[{He et~al.(2021)He, Liu, Gao, and Chen}]{he2020deberta}
Pengcheng He, Xiaodong Liu, Jianfeng Gao, and Weizhu Chen. 2021.
\newblock \href {https://openreview.net/forum?id=XPZIaotutsD} {Deberta: Decoding-enhanced bert with disentangled attention}.
\newblock In \emph{International Conference on Learning Representations}.

\bibitem[{Hu and Liu(2004)}]{hu2004mining}
Minqing Hu and Bing Liu. 2004.
\newblock Mining and summarizing customer reviews.
\newblock In \emph{Proceedings of the tenth ACM SIGKDD international conference on Knowledge discovery and data mining}, pages 168--177.

\bibitem[{Jin et~al.(2020)Jin, Jin, Zhou, and Szolovits}]{jin2020bert}
Di~Jin, Zhijing Jin, Joey~Tianyi Zhou, and Peter Szolovits. 2020.
\newblock Is bert really robust? a strong baseline for natural language attack on text classification and entailment.
\newblock In \emph{Proceedings of the AAAI conference on artificial intelligence}, volume~34, pages 8018--8025.

\bibitem[{Li et~al.(2020{\natexlab{a}})Li, Zhou, He, Wang, Yang, and Li}]{li2020sentence}
Bohan Li, Hao Zhou, Junxian He, Mingxuan Wang, Yiming Yang, and Lei Li. 2020{\natexlab{a}}.
\newblock \href {https://doi.org/10.18653/v1/2020.emnlp-main.733} {On the sentence embeddings from pre-trained language models}.
\newblock In \emph{Proceedings of the 2020 Conference on Empirical Methods in Natural Language Processing (EMNLP)}, pages 9119--9130, Online. Association for Computational Linguistics.

\bibitem[{Li et~al.(2019)Li, Ji, Du, Li, and Wang}]{li2018textbugger}
Jinfeng Li, Shouling Ji, Tianyu Du, Bo~Li, and Ting Wang. 2019.
\newblock \href {https://doi.org/10.14722/ndss.2019.23138} {{TextBugger}: Generating adversarial text against real-world applications}.
\newblock In \emph{Proceedings 2019 Network and Distributed System Security Symposium}. Internet Society.

\bibitem[{Li et~al.(2020{\natexlab{b}})Li, Ma, Guo, Xue, and Qiu}]{li_etal_2020_bert_attack}
Linyang Li, Ruotian Ma, Qipeng Guo, Xiangyang Xue, and Xipeng Qiu. 2020{\natexlab{b}}.
\newblock \href {https://doi.org/10.18653/v1/2020.emnlp-main.500} {{BERT}-{ATTACK}: Adversarial attack against {BERT} using {BERT}}.
\newblock In \emph{Proceedings of the 2020 Conference on Empirical Methods in Natural Language Processing (EMNLP)}, pages 6193--6202, Online. Association for Computational Linguistics.

\bibitem[{Li and Qiu(2021)}]{li2021token}
Linyang Li and Xipeng Qiu. 2021.
\newblock Token-aware virtual adversarial training in natural language understanding.
\newblock In \emph{Proceedings of the AAAI Conference on Artificial Intelligence}, volume~35, pages 8410--8418.

\bibitem[{Liu et~al.(2019)Liu, Ott, Goyal, Du, Joshi, Chen, Levy, Lewis, Zettlemoyer, and Stoyanov}]{liu2019roberta}
Yinhan Liu, Myle Ott, Naman Goyal, Jingfei Du, Mandar Joshi, Danqi Chen, Omer Levy, Mike Lewis, Luke Zettlemoyer, and Veselin Stoyanov. 2019.
\newblock \href {https://doi.org/10.48550/ARXIV.1907.11692} {Roberta: A robustly optimized bert pretraining approach}.

\bibitem[{Maas et~al.(2011)Maas, Daly, Pham, Huang, Ng, and Potts}]{IMDb}
Andrew~L. Maas, Raymond~E. Daly, Peter~T. Pham, Dan Huang, Andrew~Y. Ng, and Christopher Potts. 2011.
\newblock \href {http://www.aclweb.org/anthology/P11-1015} {Learning word vectors for sentiment analysis}.
\newblock In \emph{Proceedings of the 49th Annual Meeting of the Association for Computational Linguistics: Human Language Technologies}, pages 142--150, Portland, Oregon, USA. Association for Computational Linguistics.

\bibitem[{Madry et~al.(2018)Madry, Makelov, Schmidt, Tsipras, and Vladu}]{madry2017towards}
Aleksander Madry, Aleksandar Makelov, Ludwig Schmidt, Dimitris Tsipras, and Adrian Vladu. 2018.
\newblock \href {https://openreview.net/forum?id=rJzIBfZAb} {Towards deep learning models resistant to adversarial attacks}.
\newblock In \emph{6th International Conference on Learning Representations, {ICLR} 2018, Vancouver, BC, Canada, April 30 - May 3, 2018, Conference Track Proceedings}. OpenReview.net.

\bibitem[{Marelli et~al.(2014)Marelli, Menini, Baroni, Bentivogli, Bernardi, and Zamparelli}]{marelli2014sick}
Marco Marelli, Stefano Menini, Marco Baroni, Luisa Bentivogli, Raffaella Bernardi, and Roberto Zamparelli. 2014.
\newblock A sick cure for the evaluation of compositional distributional semantic models.
\newblock In \emph{Proceedings of the Ninth International Conference on Language Resources and Evaluation (LREC'14)}, pages 216--223.

\bibitem[{Miao et~al.(2021)Miao, Zhang, Xie, Song, Li, Jia, and Guo}]{miao2021simple}
Deshui Miao, Jiaqi Zhang, Wenbo Xie, Jian Song, Xin Li, Lijuan Jia, and Ning Guo. 2021.
\newblock \href {https://doi.org/10.48550/ARXIV.2111.13301} {Simple contrastive representation adversarial learning for nlp tasks}.

\bibitem[{Morris et~al.(2020)Morris, Lifland, Yoo, Grigsby, Jin, and Qi}]{morris2020textattack}
John Morris, Eli Lifland, Jin~Yong Yoo, Jake Grigsby, Di~Jin, and Yanjun Qi. 2020.
\newblock Textattack: A framework for adversarial attacks, data augmentation, and adversarial training in nlp.
\newblock In \emph{Proceedings of the 2020 Conference on Empirical Methods in Natural Language Processing: System Demonstrations}, pages 119--126.

\bibitem[{Neelakantan et~al.(2022)Neelakantan, Xu, Puri, Radford, Han, Tworek, Yuan, Tezak, Kim, Hallacy, Heidecke, Shyam, Power, Nekoul, Sastry, Krueger, Schnurr, Such, Hsu, Thompson, Khan, Sherbakov, Jang, Welinder, and Weng}]{neelakantan2022text}
Arvind Neelakantan, Tao Xu, Raul Puri, Alec Radford, Jesse~Michael Han, Jerry Tworek, Qiming Yuan, Nikolas Tezak, Jong~Wook Kim, Chris Hallacy, Johannes Heidecke, Pranav Shyam, Boris Power, Tyna~Eloundou Nekoul, Girish Sastry, Gretchen Krueger, David Schnurr, Felipe~Petroski Such, Kenny Hsu, Madeleine Thompson, Tabarak Khan, Toki Sherbakov, Joanne Jang, Peter Welinder, and Lilian Weng. 2022.
\newblock \href {https://doi.org/10.48550/ARXIV.2201.10005} {Text and code embeddings by contrastive pre-training}.

\bibitem[{Nie et~al.(2020)Nie, Williams, Dinan, Bansal, Weston, and Kiela}]{nie2019adversarial}
Yixin Nie, Adina Williams, Emily Dinan, Mohit Bansal, Jason Weston, and Douwe Kiela. 2020.
\newblock \href {https://doi.org/10.18653/v1/2020.acl-main.441} {Adversarial {NLI}: A new benchmark for natural language understanding}.
\newblock In \emph{Proceedings of the 58th Annual Meeting of the Association for Computational Linguistics}, pages 4885--4901, Online. Association for Computational Linguistics.

\bibitem[{Pan et~al.(2022)Pan, Hang, Sil, and Potdar}]{pan2022improved}
Lin Pan, Chung-Wei Hang, Avirup Sil, and Saloni Potdar. 2022.
\newblock Improved text classification via contrastive adversarial training.
\newblock In \emph{Proceedings of the AAAI Conference on Artificial Intelligence}, volume~36, pages 11130--11138.

\bibitem[{Pang and Lee(2004)}]{pang2004sentimental}
Bo~Pang and Lillian Lee. 2004.
\newblock A sentimental education: Sentiment analysis using subjectivity summarization based on minimum cuts.
\newblock In \emph{Annual Meeting of the Association for Computational Linguistics}.

\bibitem[{Pang and Lee(2005)}]{pang2005seeing}
Bo~Pang and Lillian Lee. 2005.
\newblock \href {https://doi.org/10.3115/1219840.1219855} {Seeing stars: Exploiting class relationships for sentiment categorization with respect to rating scales}.
\newblock In \emph{Proceedings of the 43rd Annual Meeting of the Association for Computational Linguistics ({ACL}{'}05)}, pages 115--124, Ann Arbor, Michigan. Association for Computational Linguistics.

\bibitem[{Pennington et~al.(2014)Pennington, Socher, and Manning}]{pennington_etal_2014_glove}
Jeffrey Pennington, Richard Socher, and Christopher Manning. 2014.
\newblock \href {https://doi.org/10.3115/v1/D14-1162} {{G}lo{V}e: Global vectors for word representation}.
\newblock In \emph{Proceedings of the 2014 Conference on Empirical Methods in Natural Language Processing ({EMNLP})}, pages 1532--1543, Doha, Qatar. Association for Computational Linguistics.

\bibitem[{Reimers and Gurevych(2019)}]{reimers2019sentence}
Nils Reimers and Iryna Gurevych. 2019.
\newblock Sentence-bert: Sentence embeddings using siamese bert-networks.
\newblock In \emph{Conference on Empirical Methods in Natural Language Processing}.

\bibitem[{Ren et~al.(2019)Ren, Deng, He, and Che}]{ren_etal_2019_generating}
Shuhuai Ren, Yihe Deng, Kun He, and Wanxiang Che. 2019.
\newblock \href {https://doi.org/10.18653/v1/P19-1103} {Generating natural language adversarial examples through probability weighted word saliency}.
\newblock In \emph{Proceedings of the 57th Annual Meeting of the Association for Computational Linguistics}, pages 1085--1097, Florence, Italy. Association for Computational Linguistics.

\bibitem[{Rima et~al.(2022)Rima, Heo, and Choi}]{rima2023adversarial}
Daniela~N. Rima, DongNyeong Heo, and Heeyoul Choi. 2022.
\newblock Adversarial training with contrastive learning in nlp.
\newblock \emph{Computer Speech \& Language}.
\newblock Submitted.

\bibitem[{Socher et~al.(2013)Socher, Perelygin, Wu, Chuang, Manning, Ng, and Potts}]{socher_etal_2013_recursive}
Richard Socher, Alex Perelygin, Jean Wu, Jason Chuang, Christopher~D. Manning, Andrew Ng, and Christopher Potts. 2013.
\newblock \href {https://aclanthology.org/D13-1170} {Recursive deep models for semantic compositionality over a sentiment treebank}.
\newblock In \emph{Proceedings of the 2013 Conference on Empirical Methods in Natural Language Processing}, pages 1631--1642, Seattle, Washington, USA. Association for Computational Linguistics.

\bibitem[{Su et~al.(2021)Su, Cao, Liu, and Ou}]{su2021whitening}
Jianlin Su, Jiarun Cao, Weijie Liu, and Yangyiwen Ou. 2021.
\newblock \href {http://arxiv.org/abs/2103.15316} {Whitening sentence representations for better semantics and faster retrieval}.
\newblock \emph{CoRR}, abs/2103.15316.

\bibitem[{Sun et~al.(2019)Sun, Qiu, Xu, and Huang}]{sun2019fine}
Chi Sun, Xipeng Qiu, Yige Xu, and Xuanjing Huang. 2019.
\newblock How to fine-tune bert for text classification?
\newblock In \emph{China national conference on Chinese computational linguistics}, pages 194--206. Springer.

\bibitem[{Wang et~al.(2021)Wang, Ding, Li, and Zheng}]{wang2021cline}
Dong Wang, Ning Ding, Piji Li, and Hai-Tao Zheng. 2021.
\newblock Cline: Contrastive learning with semantic negative examples for natural language understanding.
\newblock In \emph{Proceedings of the 59th Annual Meeting of the Association for Computational Linguistics and the 11th International Joint Conference on Natural Language Processing}, pages 2332--2342. Association for Computational Linguistics.

\bibitem[{Wang et~al.(2023)Wang, Zhang, Lei, Cao, Peng, and Wang}]{wang2023clsep}
Qian Wang, Weiqi Zhang, Tianyi Lei, Yu~Cao, Dezhong Peng, and Xu~Wang. 2023.
\newblock Clsep: Contrastive learning of sentence embedding with prompt.
\newblock \emph{Knowledge-Based Systems}, 266:110381.

\bibitem[{Wang and Isola(2020)}]{wang2020understanding}
Tongzhou Wang and Phillip Isola. 2020.
\newblock Understanding contrastive representation learning through alignment and uniformity on the hypersphere.
\newblock In \emph{International Conference on Machine Learning}, pages 9929--9939. PMLR.

\bibitem[{Wiebe et~al.(2005)Wiebe, Wilson, and Cardie}]{wiebe2005annotating}
Janyce Wiebe, Theresa Wilson, and Claire Cardie. 2005.
\newblock Annotating expressions of opinions and emotions in language.
\newblock \emph{Language resources and evaluation}, 39(2):165--210.

\bibitem[{Williams et~al.(2018)Williams, Nangia, and Bowman}]{williams2017broad}
Adina Williams, Nikita Nangia, and Samuel Bowman. 2018.
\newblock \href {https://doi.org/10.18653/v1/N18-1101} {A broad-coverage challenge corpus for sentence understanding through inference}.
\newblock In \emph{Proceedings of the 2018 Conference of the North {A}merican Chapter of the Association for Computational Linguistics: Human Language Technologies, Volume 1 (Long Papers)}, pages 1112--1122, New Orleans, Louisiana. Association for Computational Linguistics.

\bibitem[{Wu et~al.(2023)Wu, Zhang, Guo, De~Rijke, Fan, and Cheng}]{wu2023prada}
Chen Wu, Ruqing Zhang, Jiafeng Guo, Maarten De~Rijke, Yixing Fan, and Xueqi Cheng. 2023.
\newblock Prada: Practical black-box adversarial attacks against neural ranking models.
\newblock \emph{ACM Transactions on Information Systems}, 41(4):1--27.

\bibitem[{Yan et~al.(2021)Yan, Li, Wang, Zhang, Wu, and Xu}]{yan2021consert}
Yuanmeng Yan, Rumei Li, Sirui Wang, Fuzheng Zhang, Wei Wu, and Weiran Xu. 2021.
\newblock \href {https://doi.org/10.18653/v1/2021.acl-long.393} {{C}on{SERT}: A contrastive framework for self-supervised sentence representation transfer}.
\newblock In \emph{Proceedings of the 59th Annual Meeting of the Association for Computational Linguistics and the 11th International Joint Conference on Natural Language Processing (Volume 1: Long Papers)}, pages 5065--5075, Online. Association for Computational Linguistics.

\bibitem[{Yang et~al.(2019)Yang, Dai, Yang, Carbonell, Salakhutdinov, and Le}]{yang2019xlnet}
Zhilin Yang, Zihang Dai, Yiming Yang, Jaime Carbonell, Russ~R Salakhutdinov, and Quoc~V Le. 2019.
\newblock Xlnet: Generalized autoregressive pretraining for language understanding.
\newblock \emph{Advances in neural information processing systems}, 32.

\bibitem[{Zhang et~al.(2015)Zhang, Zhao, and LeCun}]{zhang2015character}
Xiang Zhang, Junbo Zhao, and Yann LeCun. 2015.
\newblock Character-level convolutional networks for text classification.
\newblock \emph{Advances in neural information processing systems}, 28:649--657.

\bibitem[{Zhang et~al.(2020)Zhang, He, Liu, Lim, and Bing}]{zhang2020unsupervised}
Yan Zhang, Ruidan He, Zuozhu Liu, Kwan~Hui Lim, and Lidong Bing. 2020.
\newblock An unsupervised sentence embedding method by mutual information maximization.
\newblock In \emph{Proceedings of the 2020 Conference on Empirical Methods in Natural Language Processing}, pages 1601--1610. Association for Computational Linguistics.

\end{thebibliography}

\appendix

\section{Training Details}
\label{Training}

we initialize our sentence encoder using the checkpoints obtained from BERT \cite{devlin_etal_2019_ber} and RoBERTa \cite{liu2019roberta}. RobustSentEmbed utilizes the representation of the [CLS] token as the starting point and incorporates a pooler layer on top of the [CLS] representations to facilitate contrastive learning objectives. The training process of RobustSentEmbed involves 4 epochs. The best checkpoint, determined by the highest average STS score, is selected for final evaluation. To train the model, we utilize a dataset consisting of $10^6$ randomly sampled sentences from English Wikipedia, as provided by the SimCSE framework \cite{gao2021simcse}. The average training time for RobustSentEmbed is 2-4 hours. As our framework is initialized with pre-trained checkpoints, it exhibits robustness that is not sensitive to batch sizes, thus enabling us to employ batch sizes of either 64 or 128.

\section{Ablation Studies}
\label{Ablation}
In this section, we conduct an analysis of the impact of five critical hyperparameters employed in the RobustSentEmbed framework on its overall performance. BERT\textsubscript{base} is employed as the encoder, and the assessment of hyperparameters is carried out using the development set of STS tasks.

\begin{figure}[t]
    \centering
    \includegraphics[width=\columnwidth]{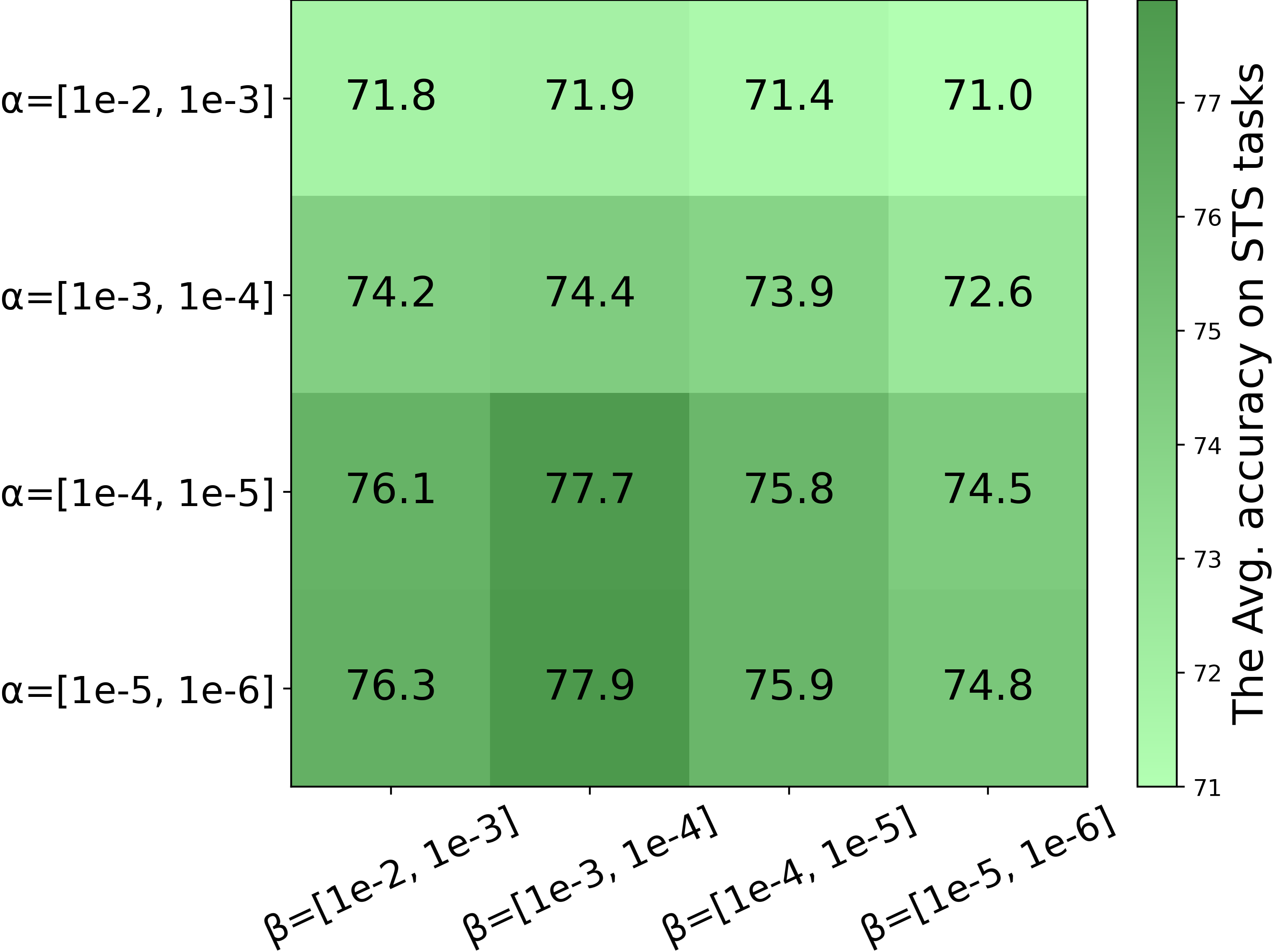}
    \caption{The impact of step sizes in perturbation generation on the average performance of STS tasks.}
    \label{fig:fig4}
\end{figure}

\subsection{Step Sizes in Perturbation Generator}

The RobustSentEmbed framework integrates two step sizes, denoted as $\alpha$ and $\beta$, to conduct iterative updates during the PGD and FGSM perturbation generation processes, respectively. Figure \ref{fig:fig4} shows the cooperative impact of adjusting the ranges for these two step sizes in generating high-risk perturbations, a crucial aspect for achieving an effective contrastive learning objective. The outcomes demonstrate more substantial improvements when $\beta$ is fine-tuned to a lower bound, coupled with $\alpha$ set to an upper bound. More precisely, enhanced performance is evident when $\alpha$ and $\beta$ are allocated ranges of [1e-4, 1e-6] and [1e-3, 1e-4], respectively. Consequently, we employ $\alpha$ = 1e-5 and $\beta$ = 1e-3 for our experiments, as this configuration yields the optimal results among the different configurations.

\subsection{Step Numbers in Perturbation Generator}
\begin{figure}[t]
    \centering
    \includegraphics[width=\columnwidth]{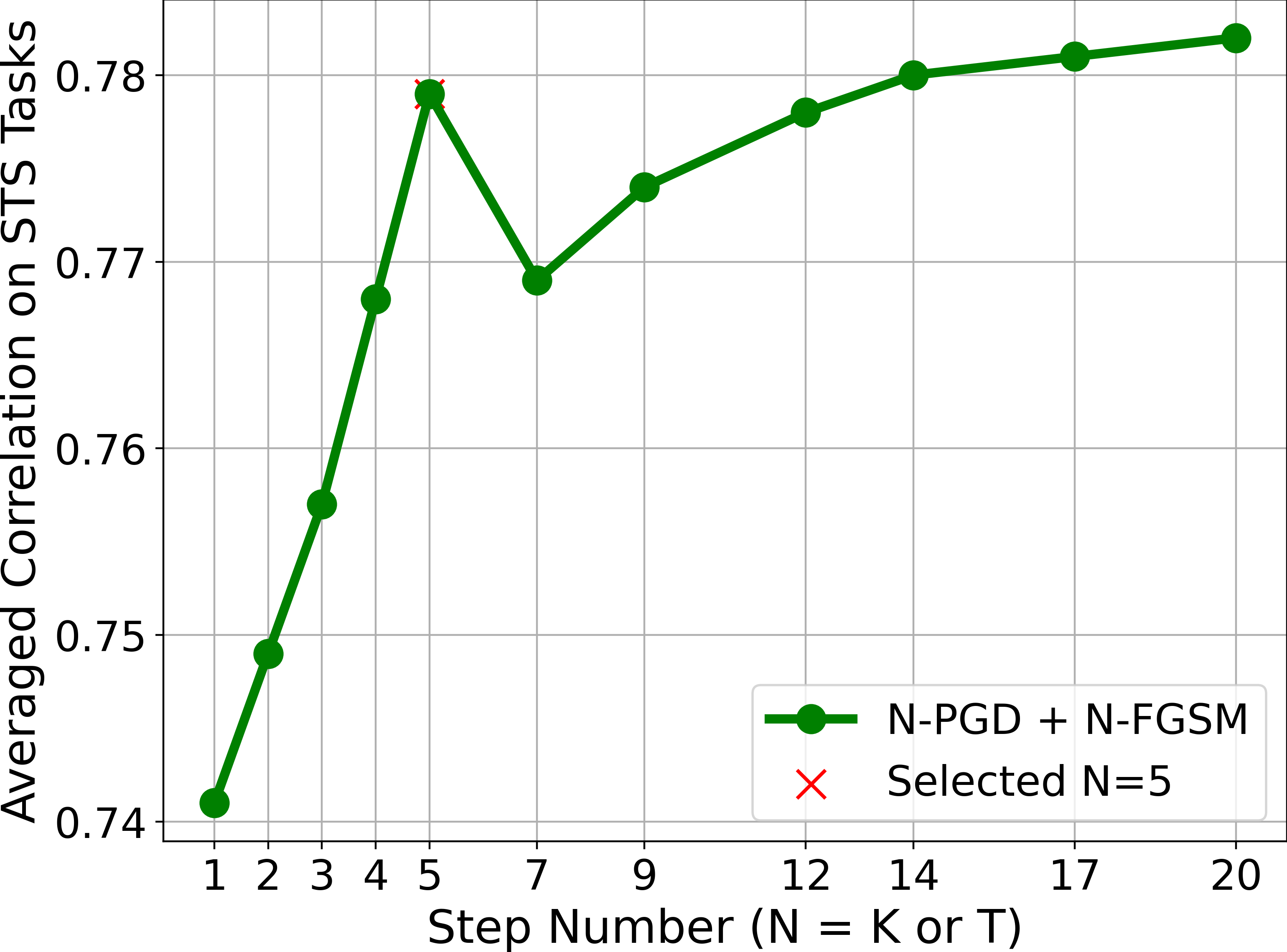}
    \caption{The impact of the step number (represented by N = K or T) in the T-step FGSM and K-step PGD methods on the averaged correlation of the STS tasks.}
    \label{fig:fig5}
\end{figure}

RobustSentEmbed employs T-step FGSM and K-step PGD iterations to acquire high-risk adversarial perturbations for the contrastive learning objective. For simplicity in perturbation generation analysis, we establish K = T. The influence of varying step numbers (N = K or T) on effectiveness is illustrated in Figure \ref{fig:fig5}. A gradual improvement is observed as N increases from 1 to 12; however, beyond N=12, the improvement becomes negligible. Additionally, higher N results in longer running time and inequitable resource allocation. Consequently, we opt for N=5 in our experiments.  

\subsection{Norm Constraint}

To ensure imperceptibility in the generated adversarial examples, RobustSentEmbed regulates the magnitude of the perturbation vectors (whether $\boldsymbol{\delta}$ or $\boldsymbol{\eta}$). This control is achieved through the utilization of three commonly employed norm functions: $L_1$, $L_2$, and $L_\infty$, to  restrict the magnitude of the perturbation to small values. The averaged Spearman’s correlation of these norm functions across different Semantic Textual Similarity tasks is presented in Table \ref{tab:tbl8}. The $L_\infty$ norm exhibits superior correlation in comparison to the other two norms, thus warranting its selection as the norm function for our experimental assessment.

\begin{table}[htbp]
  \centering
  \begin{tabular}{lc}
\hline
    
 \bfseries Norm & \bfseries Correlation \\
    \hline
    \hline
    
    $L_\infty$ & \textbf{77.90} \\
    $L2$ & 76.84 \\
    $L1$ & 76.52 \\
    
\hline
  \end{tabular}
\caption{The impact of the norm constraint on perturbation generation on the average performance of various STS tasks.}
\label{tab:tbl8}
\end{table}

\subsection{Contrastive Learning Loss}

\begin{figure}[t]
    \centering
    \includegraphics[width=\columnwidth]{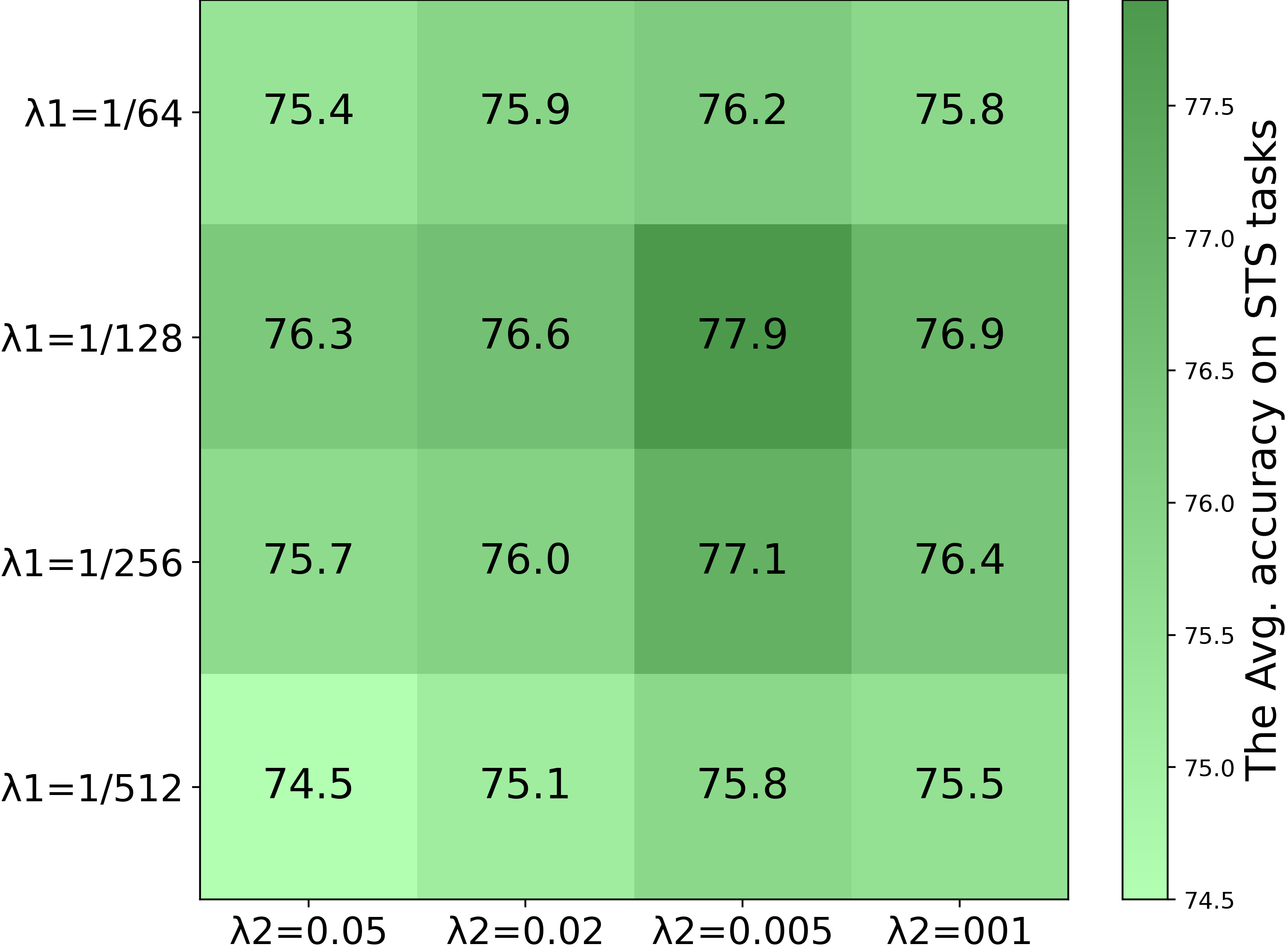}
    \caption{The impact of weighting coefficients in the total loss function on the average performance of STS tasks.}
    \label{fig:fig6}
\end{figure}

The first part of the total loss function (Equation \ref{total_loss}) is dedicated to optimizing the similarity between the input instance $x$ and its positive pair ($x^{pos}$), as well as the similarity between $x$ and its adversarial perturbation ($x^{adv}$). While this indirectly brings $x^{pos}$ and $x^{adv}$ closer, our findings indicate that incorporating direct contrastive learning between $x^{pos}$ and $x^{adv}$ (the second part of Equation \ref{total_loss}) through the regularization of the objective function in the first part helps us achieve enhanced clean accuracy and robustness. Additionally, the third part of the total loss function introduces the adversarial replaced token detection objective into the loss function, making it more challenging for adversarial training to converge. Figure \ref{fig:fig6} illustrates the impact of different values of the weighting coefficients (i.e., $\lambda_1$, $\lambda_2$) on the final performance of our framework. As illustrated, when $\lambda_1=1/128$ and $\lambda_2=0.005$, the framework achieves the highest average accuracy for semantic textual similarity tasks. We utilize $\lambda_1=1/128$ and $\lambda_2=0.005$ for all other experiments.
 
\subsection{Modulation Factor}

RobustSentEmbed includes a modulation factor, represented as $0\le\rho\le1$, to adjust the relative importance of each individual perturbation (PGD and FGSM) in the formation of the sentence-level perturbation. The efficacy of different values of this modulation factor on semantic textual similarity tasks is detailed in Table \ref{tab:tbl6}. The findings reveal that $\rho = 0.5$ yields the highest averaged correlation across the examined magnitudes, underscoring its capability to generate more powerful perturbations. Consequently, we employ this configuration in the setup of our framework.

\begin{table}[htbp]
  \centering
  \begin{tabular}{lc}
\hline
    
    $\boldsymbol{\rho}$ & \bfseries Correlation \\
    \hline
    \hline
    
    0 & 76.06 \\
    0.25 & 76.85 \\
    0.5 & \textbf{77.90} \\
    0.75 & 77.34 \\
    1 & 76.34 \\
    
\hline
  \end{tabular}
\caption{The impact of the modulation factor on the average performance of different Semantic Textual Similarity (STS) tasks in generating the final perturbation.}
\label{tab:tbl6}
\end{table}

\section{Adversarial Attack Methods}
\label{attacks} 

This section provides additional details regarding the various adversarial attacks. The TextBugger method \cite{li2018textbugger} identifies crucial words by analyzing the Jacobian matrix of the target model and selects the optimal perturbation from a set of five generated perturbations. The PWWS \cite{ren_etal_2019_generating} employs a synonym-swap technique based on a combination of word saliency scores and maximum word-swap effectiveness. TextFooler \cite{jin2020bert} identifies significant words, gathers synonyms, and replaces each such word with the most semantically similar and grammatically correct synonym. The BAE \cite{garg2020bae} employs four adversarial attack strategies involving word replacement and/or word insertion operations to generate substitutions. The BERTAttack \cite{li_etal_2020_bert_attack} comprises two steps: (a) identifying vulnerable words/sub-words and (b) utilizing BERT MLM to generate semantic-preserving substitutes for the vulnerable tokens.

\section{RobustSentEmbed Algorithm}

\label{algorithm} 

 Algorithm \ref{algorithm1} illustrates our framework's approach to generating a norm-bounded perturbation at both the token-level and sentence-level using an iterative process. It confuses the $f_\theta(\cdot)$ encoder by treating the perturbed embeddings as different instances. Our framework then utilizes a contrastive learning objective in conjunction with a replaced token detection objective to maximize the similarity between the embedding of the input sentence and the adversarial embedding of its positive pair (former objective), as well as its edited sentence (latter objective).

\begin{algorithm}[!ht]
\caption{RobustSentEmbed Algorithm}
\label{algorithm1}
\begin{small}

    \KwInput{Epoch number $E$, PLM Encoder $\boldsymbol{\mathbf{f_\theta}}$, dataset of raw sentences $\boldsymbol{\mathrm{\mathcal{D}}}$ , embedding perturbation \{$\boldsymbol{\delta}$, $\boldsymbol{\eta}$\}, dropout masks $m_1$ and $m_2$, perturbation bound $\epsilon$, adversarial step sizes \{$\alpha$, $\beta$, $\gamma$\}, learning rate $\xi$, perturbation modulator $\rho$, weighting coefficients \{$\lambda_1$, $\lambda_2$\}, adversarial steps \{$K$, $T$\}, contrastive learning objective ${\mathcal{L}}_{con,\theta}$ (eq. \ref{CLobjective}), ELECTRA generator $G(.)$ and discriminator $D(.)$}
    
    \KwOutput{Robust Sentence Representation} 
  
    $\mathcal{V} \in \mathbb{R}^{N*D} \gets \frac{1}{\sqrt{D}} \mathrm{U}(-\sigma, \sigma)$ 
    
    \For{$\mathrm{epoch} = 1, ..., E$}
    {
        \For{minibatch $\mathrm{B} \subset \boldsymbol{\mathrm{\mathcal{D}}}$}
        {
             $\boldsymbol{\delta}^0 \gets \frac{1}{\sqrt{D}} \mathrm{U}(-\sigma, \sigma) \: , \: \boldsymbol{\eta}^{0}_{i}  \gets \mathcal{V}[w_i] $\\

            $\boldsymbol{X} \:\:\:= \boldsymbol{\mathbf{f_\theta}}.$embedding($\mathrm{B},\:m_1$)\\
             $\boldsymbol{X^+} = \boldsymbol{\mathbf{f_\theta}}.$embedding($\mathrm{B},\:m_2$)\\
            \For{t = $1, ..., max(K,\;T)$}
              {
                
                \begin{math}
                  \boldsymbol{g}_{\delta}= \nabla_{\delta}{\mathcal{L}}_{con,\theta}(\boldsymbol{X} +\boldsymbol{\delta}^{t-1} +\boldsymbol{\eta}^{t-1}, \{\boldsymbol{X}^{+}\})
                \end{math}
                
                \If{$t \le K$}{
                $\boldsymbol{\delta}^{t}_{pgd}=\Pi_{\lVert \boldsymbol{\delta} \rVert_{P}\leq\epsilon}(\boldsymbol{\delta}^{t-1} + \alpha g(\boldsymbol{\delta}^{t-1})/\lVert g(\boldsymbol{\delta}^{t-1}) \rVert_{P})$
                }

                \If{$t \le T$}{
                $\boldsymbol{\delta}^{t}_{fgsm}=\Pi_{\lVert \boldsymbol{\delta} \rVert_{P}\leq\epsilon}(\boldsymbol{\delta}^{t-1} + \beta \mathrm{sign}(g(\boldsymbol{\delta}^{t-1}))) $
                  }

                \begin{math}
                  \boldsymbol{g}_{{\eta}_{i}}= \nabla_{\eta}{\mathcal{L}}_{con,\theta}(\boldsymbol{X} +\boldsymbol{\delta}^{t-1} +\boldsymbol{\eta}^{t-1}, \{\boldsymbol{X}^{+}\})
                \end{math}

                \begin{math}
                \boldsymbol{\eta}^{t}_{i}= n^i*(\boldsymbol{\eta}^{t}_{i-1} + \gamma \boldsymbol{g}_{{\eta}_{i}}/\lVert \boldsymbol{g}_{{\eta}_{i}} \rVert_{P})
                \end{math}

                \begin{math}
                \boldsymbol{\eta}^{t} \gets \Pi_{\lVert \boldsymbol{\eta} \rVert_{P}\leq\epsilon} (\boldsymbol{\eta}^{t})
                \end{math}

              }

              \begin{math}
              \mathcal{V}[w_i] \gets \boldsymbol{\eta}^{max(K, \:T)}_{i}
              \end{math}
              
               \begin{math}
                  \boldsymbol{\delta}_{f}= \rho\boldsymbol{\delta}^{K}_{pgd} +(1-\rho)\boldsymbol{\delta}^{T}_{fgsm}
                  \label{final_pert}
                \end{math}
               
            \For{x $ \in \mathrm{B}$}
            {
                $x^{''} = G(\mathrm{MLM}(x))$ \\

                $X^{adv}= {X}^{''} + \boldsymbol{\eta}^{max(K, \:T)}_{i}$ \\

            \scalebox{0.67}{  
            \begin{math}
                 {\mathcal{L}}^{x}_{RTD,\,\theta} = \sum_{j=1}^{\lvert x \rvert} [\mathds{-1} (X^{adv}_{j} = {X}_j) \; \mathrm{log} \: D (X^{adv},\, \boldsymbol{\mathbf{f_\theta}}(x),\, j)
           \end{math} }
           
            \scalebox{0.67}{  
            \begin{math}
                 \mathds{-1} (X^{adv}_{j} \neq {X}_j) \; \mathrm{log} \: (1-D (X^{adv},\, \boldsymbol{\mathbf{f_\theta}}(x),\, j))]
           \end{math} }
           
            }

            ${\mathcal{L}}_{RTD,\,\theta}= \sum_{i=1}^{\lvert B \rvert} {\mathcal{L}}^{x_i}_{RTD} $

             \begin{math}
                \label{eqn:CLobjective}
                {\mathcal{L}}_{RobustEmbed,\,\theta} := {\mathcal{L}}_{con,\,\theta}(\boldsymbol{X}, \{\boldsymbol{X^+},\;\boldsymbol{X}+\boldsymbol{\delta}_{f}\})
             \end{math}
\begin{align*}
    \scalebox{0.85}{$
        \begin{aligned}
            \mathcal{L}_{\text{total}} &:= \mathcal{L}_{\text{RobustEmbed},\theta}  + \lambda_1 \cdot \mathcal{L}_{\text{con},\theta}(\boldsymbol{X}+\boldsymbol{\delta}_{f}, \{\boldsymbol{X^+}\}) \\
            &\quad + \lambda_2 \cdot \mathcal{L}_{\text{RTD},\theta}
        \end{aligned}
    $}
\end{align*}

            $\theta = \theta - \xi\nabla_{\boldsymbol\theta}{\mathcal{L}}_{total}$
               
        }
    }
\end{small}
\end{algorithm}

\end{document}